\pdfoutput=1
\documentclass[11pt]{article}

\usepackage[preprint]{acl}

\usepackage{times}
\usepackage{latexsym}

\usepackage[T1]{fontenc}
\usepackage[utf8]{inputenc}

\usepackage{microtype}

\usepackage{inconsolata}

\usepackage{graphicx}
\usepackage{lipsum}
\usepackage{tabularx}
\usepackage{booktabs}
\usepackage{makecell}
\usepackage{multirow}
\usepackage{multicol}
\usepackage{bbding}
\usepackage{colortbl}
\usepackage{xcolor}
\usepackage{underscore}
\usepackage[normalem]{ulem}
\usepackage{latexsym}
\title{CFBench: A Comprehensive Constraints-Following Benchmark for LLMs}

\author{
  \textbf{Tao Zhang}\textsuperscript{1}\thanks{\ Equal contribution.}$^{\clubsuit}$ \quad
  \textbf{Chenglin Zhu}\textsuperscript{2}\footnotemark[1] \quad
  \textbf{Yanjun Shen}\textsuperscript{1}\footnotemark[1] \quad
  \textbf{Wenjing Luo}\textsuperscript{1} \quad
  \textbf{Yan Zhang}\textsuperscript{1} \\
  \textbf{Hao Liang}\textsuperscript{2} \quad
  \textbf{Tao Zhang}\textsuperscript{1}$^{\diamondsuit}$ \quad
  \textbf{Fan Yang}\textsuperscript{1} \quad
  \textbf{Mingan Lin}\textsuperscript{1} \quad
  \textbf{Yujing Qiao}\textsuperscript{1} \\
  \textbf{Weipeng Chen}\textsuperscript{1} \quad
  \textbf{Bin Cui}\textsuperscript{2} \quad
  \textbf{Wentao Zhang}\textsuperscript{2}\thanks{\ Corresponding author.} \quad
  \textbf{Zenan Zhou}\textsuperscript{1}\footnotemark[2] \\
  \textsuperscript{1}Baichuan Inc. \quad
  \textsuperscript{2}Peking University \\
  \texttt{\{zhangtao2, shenyanjun, zhouzenan\}@baichuan-inc.com} \\
  \texttt{zhuchenglin@stu.pku.edu.cn,\quad wentao.zhang@pku.edu.cn  } 
}

\begin{document}
\maketitle
\begin{abstract}
The adeptness of Large Language Models (LLMs) in comprehending and following natural language instructions is critical for their deployment in sophisticated real-world applications. Existing evaluations mainly focus on fragmented constraints or narrow scenarios, but they overlook the comprehensiveness and authenticity of constraints from the user's perspective. To bridge this gap, we propose CFBench, a large-scale Chinese Comprehensive Constraints Following Benchmark for LLMs, featuring 1,000 curated samples that cover more than 200 real-life scenarios and over 50 NLP tasks. CFBench meticulously compiles constraints from real-world instructions and constructs an innovative systematic framework for constraint types, which includes 10 primary categories and over 25 subcategories, and ensures each constraint is seamlessly integrated within the instructions. To make certain that the evaluation of LLM outputs aligns with user perceptions, we propose an advanced methodology that integrates multi-dimensional assessment criteria with requirement prioritization, covering various perspectives of constraints, instructions, and requirement fulfillment. Evaluating current leading LLMs on CFBench reveals substantial room for improvement in constraints following, and we further investigate influencing factors and enhancement strategies. The data and code will be made available.  
\end{abstract}

\section{Introduction}
% CFBench Case展示示意图(introduction)
\begin{figure}[t]
\centering
\includegraphics[width=0.95\columnwidth]{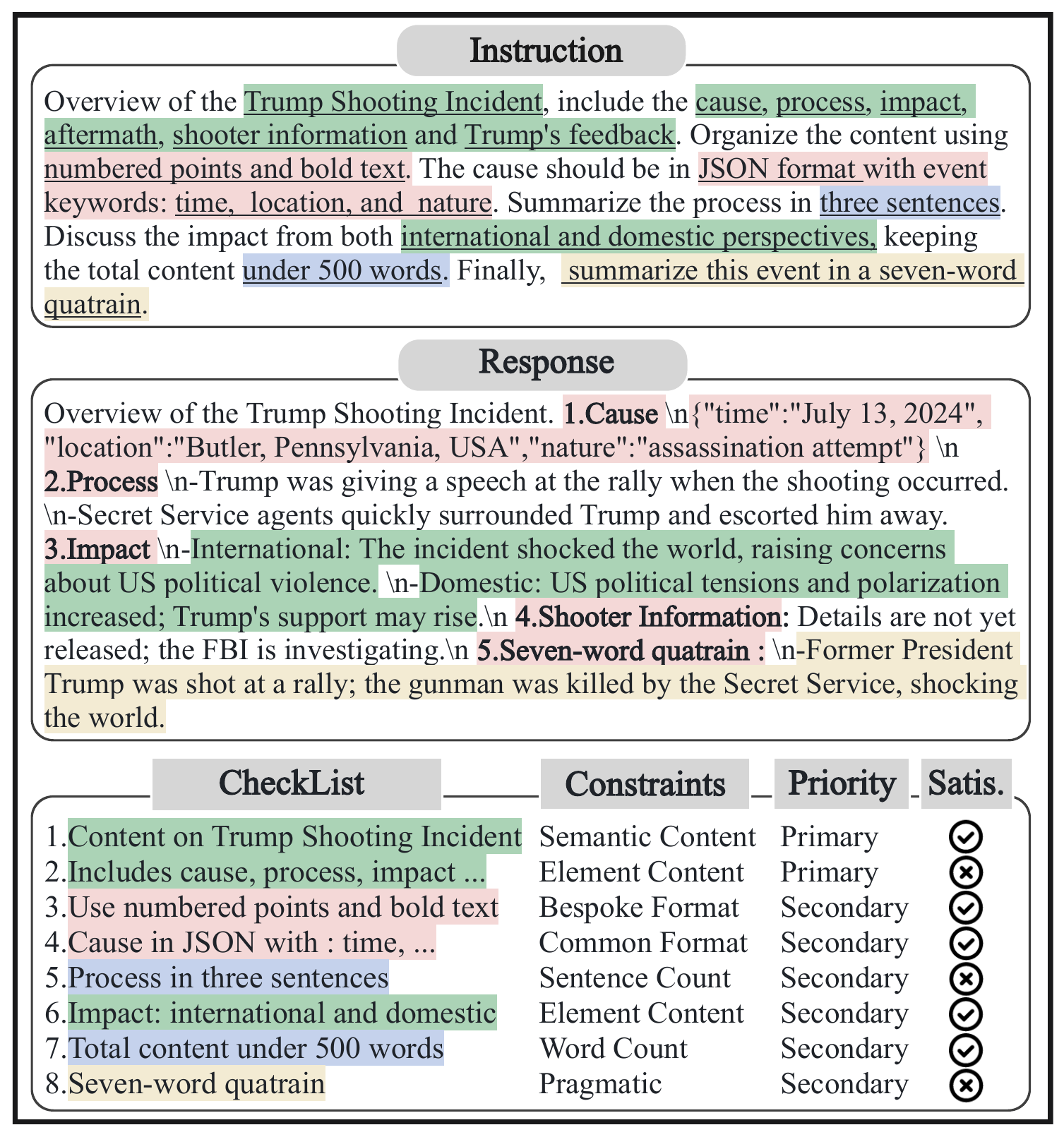}  
\caption{Sample data from CFBench. A checklist, constraint type, requirement priority, and satisfaction constitute our evaluation criteria.} 
\label{CFBench Example} 
\vspace{-20pt} % 调整这个值来减少空白
\end{figure}

Large Language Models (LLMs) have become the cornerstone of numerous cutting-edge research tasks and are widely utilized in real-world scenarios~\citep{brown2020language,chowdhery2023palm,achiam2023gpt,touvron2023llama}. In real-world scenarios, human instructions are inherently complex and accompanied by explicit constraints, requiring models to both understand intricate requirements and strictly comply with these constraints~\cite{yang2023foundation,zhong2021adapting,mishra2022cross,wei2021finetuned,sanh2022multitask}. The proficiency of LLMs in comprehending requirements and adhering to natural language constraints is essential, as it ensures tasks are executed precisely and resolved perfectly according to user instructions.

The prevailing method for evaluating a model's instruction-following ability involves using quantitative programs, human evaluators, or advanced LLMs to assess response quality across single constraints, complex problems, and finite constraints ~\cite{zhou2023instruction,wang2023self,li2023alpacaeval,zheng2024judging,xu2023wizardlm}. ~\citet{laskar2024systematic} underscores the importance of evaluating data quality, highlighting the necessity for real and extensive data distribution, along with its applicability to real-world scenarios. ~\citet{sun2024comprehensive} also stresses that realistic evaluation metrics reflect model capabilities and guide iteration. Constraints-following evaluation faces analogous challenges, particularly within complex real-world scenarios, where data sources and contexts are diverse, and where evaluation is both subjective and arduous. Fig. \ref{CFBench Example}, which addresses the aforementioned challenges, presents a sample from CFBench illustrating the Trump assassination event with different colors representing various constraint types. The instruction include multiple constraints, and the evaluation method uses a checklist to break down complex requirements into independent checkpoints, annotating constraint types and priorities. LLMs are then used to assess each checkpoint. For the English-Chinese comparison example, see Appendix Fig.~\ref{CFBench Case: Chinese and English Comparison}. Consequently, we introduce two more profound challenges in constraints-following assessment.

\textbf{Q1: How to construct high-quality evaluation data?}
Many studies focus on evaluating single constraint~\cite{chen2022controllable,tang2023struc}, lacking comprehensive analysis across diverse constraints.~\citet{he2024can} examines LLM performance on complex real-world instructions but neglect constraint diversity and scenario coverage.~\citet{jiang2023followbench} incrementally incorporate fine-grained constraints to craft multi-level instructions. However, with only 75 instances of mixed type, which risks variability due to limited data, and equating difficulty with constraint quantity oversimplifies the task. Recent work focuses on evaluating constraints combinability~\cite{wen2024benchmarking}. To ensure data quality, we systematically categorize constraints by mining real-world online data and using classification, synthesis, and expert design, covering 10 primary categories and over 25 subcategories. We also cross-match these constraints with various domains and scenarios, ensuring balanced representation and expert-validated reasonableness.   

\textbf{Q2: How to evaluate accurately and meticulously?}
Evaluating LLMs' adherence to constraints is challenging and typically involves manual, automated, and programmatic assessments using various metrics. Representative work computes outcomes for verifiable instructions using code ~\cite{zhou2023instruction,he2024can}. 
~\citet{jiang2023followbench} uses scripts and constraint-evolution paths to handle diverse challenging instructions, introducing three metrics tailored to the data's characteristics. The DRFR method decomposes complex constraints into binary judgments, with GPT evaluating each criterion~\cite{qin2024infobench}. Indeed, previous work has ensured the feasibility and objectivity of evaluations through various methods, but they have overlooked assessments from the user's multiple perspectives. We deconstruct complex instructions from the user's perspective into multiple sub-needs, categorizing them by priority and constraint type, with LLMs evaluating each checkpoint. Furthermore, a multi-dimensional evaluation criteria is proposed using three metrics from the perspectives of constraints, instructions, and requirements priority.

%To address the two challenges
We introduce CFBench, a comprehensive Chinese benchmark designed to thoroughly evaluate the constraint comprehension and following capabilities of LLMs. CFBench comprises 10 primary categories and over 25 secondary subcategories organized through taxonomic and statistical methodologies. CFBench features 1,000 meticulously curated samples spanning more than 200 real-life scenarios across 20 domains and over 50 NLP tasks, enhancing the breadth and generality of the evaluation data. Additionally, we have seamlessly integrated original instructions and constraint types within each sample, paying particular attention to nuanced combinations, ensuring each constraint is credibly and coherently embedded. 
% Our advanced evaluation methodology incorporates multi-dimensional assessment criteria, which prioritizing requirements to align LLM outputs with user needs, enhance interpretability, and facilitate iterative development. 
Finally, extensive experiments and exploratory discussions provide strong support for evaluation and optimization.

Overall, our contributions are mainly four-fold:

\begin{itemize}
\item To the best of our knowledge, we are the pioneers in systematically defining an instruction constraint framework utilizing both taxonomic and statistical methodologies.
\item We introduce CFBench, a meticulously annotated, large-scale, high-quality Chinese benchmark that encompasses a broad spectrum of real-world scenarios and NLP tasks.
\item We propose a multi-dimensional evaluation framework to comprehensively assess model capabilities while prioritizing user-centric needs.
\item We exhaustively evaluated prominent LLMs, uncovering significant deficiencies in constraints following and exploring performance factors and optimization strategies. 

\end{itemize}

% CFBench数据构建&约束构建流程图
\begin{figure*}[t]
\centering
\includegraphics[width=0.95\textwidth]{cfbench_fig/case2_2.pdf} % Reduce the figure size so that it is slightly narrower than the column.
\caption{The construction pipeline and evaluation sample of CFBench. Initially, it entails the construction of the constraint system, followed by the assembly of the dataset, and culminating in the proposal of a multi-perspective user view evaluation.} 
\label{CFBench Construction Pipeline}
\end{figure*}

\section{Related Work}
\iffalse
\subsection{Evaluation for LLMs} 
Numerous studies have concurrently evaluated LLMs from various perspectives. By integrating a multitude of existing datasets, many have assessed the overall capabilities of LLMs ~\cite{bommasani2023holistic,zhong2024agieval,dubois2024alpacafarm,chia2024instructeval,hendrycks2021ethics}. Some research has delved into specialized capabilities, such as programming ~\cite{chen2021evaluating}, reasoning ~\cite{cobbe2021training}, and knowledge ~\cite{huang2024c}. Unlike previous studies, we evaluate LLMs' instruction-following abilities with a focus on constraint-based instructions.  
\fi

\subsection{Instruction Following}

Fine-tuning LLMs with annotated instructional data enhances their ability to follow general language instructions ~\cite{weller2020learning,sanh2022multitask}. Studies show that more complex or constrained instructions further improve this ability. For instance, six methods to create intricate instructions from a small set of handwritten seed data are proposed \cite{xu2023wizardlm}, while ~\citet{mukherjee2023orca} elevate training data complexity by having GPT-4\cite{achiam2023gpt} generate reasoning steps for simple instructions. The latest work ~\cite{sun2024conifer,he2024complex,dong2024self} suggests that increasing the number and variety of constraints can enhance the complexity of instructions, thereby further improving the model's ability to follow constraint-based instructions. 

\subsection{Evaluation of Constraints Following}
Constraints such as word count, position, topics, and content have garnered significant attention in the field of Controlled Text Generation ~\cite{yao2023collie,zhou2023controlled}. 
\citet{zhou2023instruction} centers on assessing 25 verifiable instructions. Numerous studies have explored the adherence of LLMs to format constraints, such as complex tabular data ~\cite{tang2023struc} and customized scenario formats ~\cite{xia2024fofo}. ~\citet{qin2024infobench} decomposing a single instruction into multiple constraints. \citet{he2024can} gathered constraints from real-world scenarios and developed a sophisticated benchmark using detailed task descriptions and inputs.~\citet{jiang2023followbench} progressively integrates fine-grained constraints to develop multi-level instructions, thereby enhancing complexity across six distinct types. Concurrent work ~\cite{wen2024benchmarking}, constructs a novel benchmark by synthesizing and refining data from the aforementioned benchmarks, with an emphasis on the combinatorial types of constraints. However, previous studies suffered from fragmented constraints, limited scenarios, and misaligned evaluation methods with user perspectives.

% % CFBench数据集整体统计表
\newcolumntype{C}{>{\centering\arraybackslash}X}
\begin{table*}
\centering
\small % 调整字体大小
\begin{tabularx}{\textwidth}{p{1.25cm}CCCCC|CCCCCCCCCC}
\toprule[0.1em]
\multirow{2}*{\textbf{Split Set}} & \multicolumn{5}{c}{\textbf{Basic Info}} & \multicolumn{10}{c}{\textbf{Constraints Count}} \\
\cmidrule(lr){2-6} \cmidrule(lr){7-16}
~&\textbf{Num.}&\textbf{Len.}&\textbf{Prim.}&\textbf{Cons.}&\textbf{Type.}   &\textbf{C1}&\textbf{C2}&\textbf{C3} &\textbf{C4} &\textbf{C5} &\textbf{C6} &\textbf{C7} &\textbf{C8} &\textbf{C9} &\textbf{C10} \\
\midrule
Easy Set   &500&413&1.69&3.59&2.83	   &613&214&180&170&134&92&82&95&90&79  \\
Hard Set   &500&605&1.98&4.89&3.58	   &772&345&233&241&168&122&115&145&137&81  \\
\midrule
Full Set    &1000&509&1.84&4.24&3.20	   &1385&559&413&411&302&214&197&240&227&160  \\
\bottomrule[0.1em]
\end{tabularx}
\caption{\label{CFBench overall Statistic} CFBench Statistic. The abbreviations of 'Num.', 'Len.', 'Prim.', 'Cons.', 'Type.' denote the sample number, average instruction length, average primary requirements number, average constraint number, average constraint type number, respectively. The designations ’C1’-’C10’ denote the Primary Constraint types of content, numerical, style, format, linguistic, situation, example, inverse, contradictory, and rule constraint, respectively.}
\end{table*}

% CFBench数据集整体统计表

\section{CFBench}
As depicted in Fig. \ref{CFBench Construction Pipeline}, the CFBench construction pipeline includes several key components. First, we collect and systematize constraint expressions from real-world scenarios and various NLP tasks. Using this system, we create high-quality evaluation data by combining instructions from these scenarios with advanced LLMs and manual curation. We then introduce innovative multi-perspective evaluation method.
Additionally, we conduct a thorough statistical analysis and validate the quality from various angles to highlight reliability and applicability.

\subsection{Constraints System}
\subsubsection{Constraints Collection}
We amass a diverse corpus of instructions from real-world scenarios and various NLP tasks ~\cite{xia2024fofo,li2024cif} to ensure a comprehensive system. Initially, we aggregate several million instructions from online logs and NLP tasks, refining these through length filtering and clustering to distill 30,000 high-quality instructions. Utilizing advanced LLM techniques, we extract and expand atomic constraints through evolutionary methods. Using LLMs, we carefully select meaningful atomic constraints, resulting in over 5000 unique constraints.  
Domain experts first filter out unreasonable or meaningless atomic constraints and then synthesize these into a structured framework with 10 primary categories and 25 subcategories, guided by principles of statistics, taxonomy, and linguistics.

\subsubsection{Constraints System} 
\textbf{Content constraints} control the scope and depth of output content by specifying certain conditions~\cite{zhang2023survey}, and can be divided into lexical constraints, element constraints, and semantic constraints based on their granularity. 
\textbf{Numerical constraints}, which ensure that output content meets length and quantity requirements~\cite{yao2023collie}, can be classified into word-level, sentence-level, paragraph-level, document-level based on the objects involved in the planning.
\textbf{Stylistic constraints} impart a unique flavor and color to the output, revealing the author's traits and chosen social objectives~\cite{tsai2021style}, can be subdivided into tonal, formal, audience, and authorial style constraints based on the perspective of application.
\textbf{Format constraints}~\cite{tang2023struc} standardize expression to guide the generation of complex content and can be categorized into fundamental, bespoke, and specialized scenario constraints based on their usage scenarios.  
\textbf{Linguistic constraints}~\cite{zhou2023controlled} adapt to various scenarios by controlling internal features and logic, grouped into Pragmatic, Syntactic, Morphological, Phonological, and other constraints. 
\textbf{Situation constraints}~\cite{liu2023agentbench} guide response appropriateness through background or situational parameters, can be classified into role-based, task-specific, and complex contextual constraints.
\textbf{Example constraints} regulate new responses by leveraging the intrinsic patterns established by a limited set of samples, with an emphasis on assessing the model's proficiency in contextual constraint learning. 
\textbf{Inverse constraints} narrow the response space through the mechanism of indirect exclusion.  
\textbf{Contradictory constraints} denote conditions that are mutually exclusive, rendering it impossible for the response to fulfill all requirements concurrently, which are prevalent in online logs and are often easily overlooked. 
\textbf{Rule constraints} define logic flows or actions and meticulously crafted to standardize the road of responses. Details are in Appendix Tab.~\ref{constraint system}.
% Detailed definitions, and examples can be found in Appendix Tab.~\ref{constraint system}.
% classifications, of the constraint system 

\subsection{Dataset Construction}
To guarantee data quality in terms of authority and thorough coverage, we utilize a collaborative iterative methodology that synergizes expertise with the capabilities of LLMs.

\subsubsection{Data Source and Selection}
Real-world scenarios and NLP tasks form the foundation for CFBench's initial instructions. By harnessing advanced LLMs, we rigorously assess each instruction for constraint types and quantities within a predefined system, filtering out those with unreasonable or ineffective constraints. Subsequently, we balance the scenarios and constraint types, resulting in a refined set of 2,000 instructions covering all scenarios and NLP tasks. Prompts and checklist generation are in Appendix. 

\subsubsection{Iterative Refinement}
Professional annotators carefully review and refine the data, ensuring the rationality of constraints and gold answer. If modifications are needed, instructions are revised, and LLMs generate responses with refined evaluation criteria, repeating this process until satisfactory results are achieved. Ultimately, comprehensive support is formulated for each sample, detailing high-quality instructions, the ideal answer, specific assessment criteria, constraint types, and priority levels. 

\begin{figure}[t]
\centering
\includegraphics[width=0.9\columnwidth]{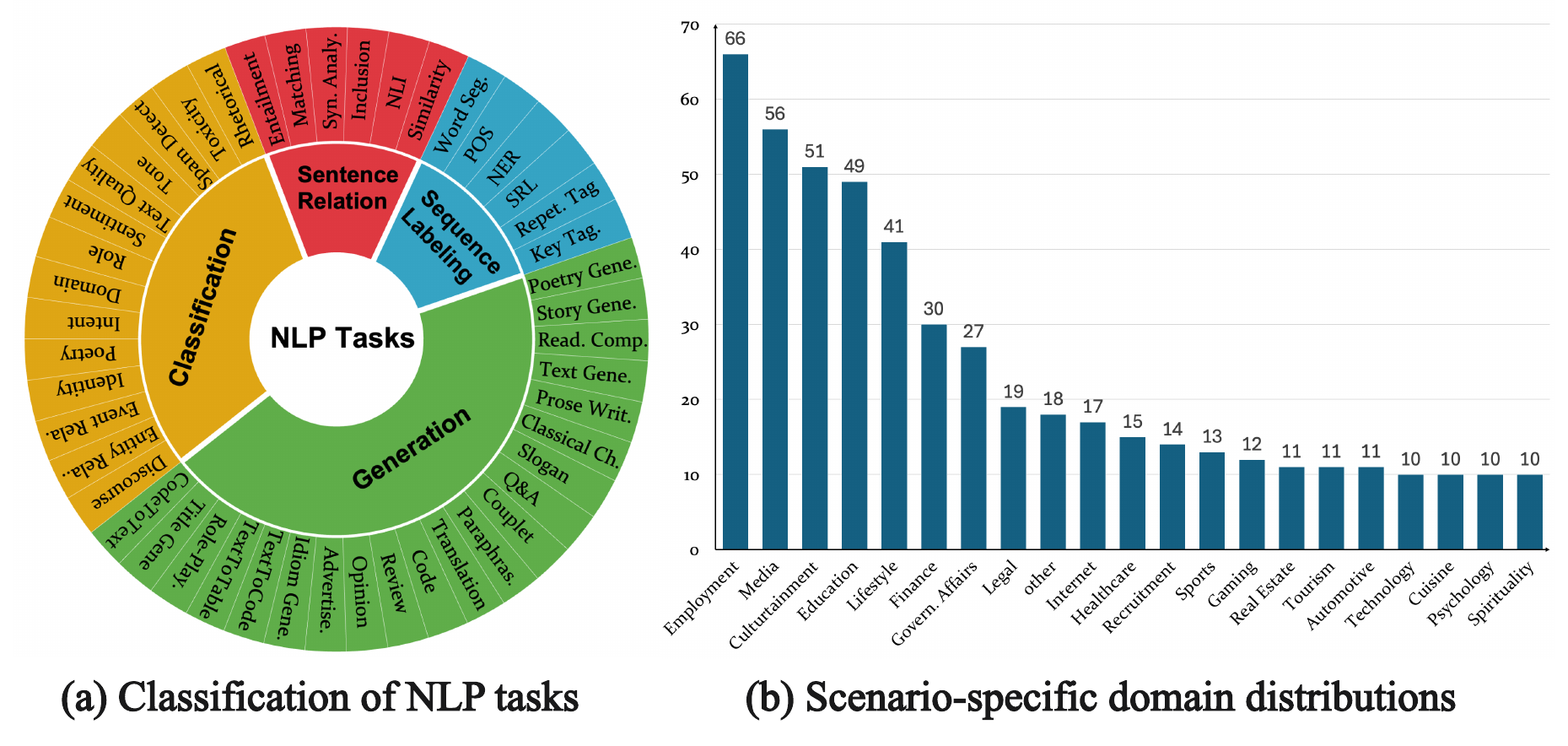} % Reduce the figure size so that it is slightly narrower than the column. Don't use precise values for figure width.This setup will avoid overfull boxes.
\caption{The distribution of NLP tasks and domains} 
\label{Distribution of Domain and Scenario}
\end{figure}

% %实验结果：
% \newcolumntype{C}{>{\centering\arraybackslash}X}
\begin{table*}
\centering
\small % 调整字体大小
\begin{tabularx}{\textwidth}{p{3.8cm}CCC|CCC|CCC}
\toprule[0.1em]
    \multirow{2}*{\textbf{Models}} & \multicolumn{3}{c}{\textbf{Easy Set}} &\multicolumn{3}{c}{\textbf{Hard Set}} &\multicolumn{3}{c}{\textbf{Full Set}}\\    
    \cmidrule(lr){2-4} \cmidrule(lr){5-7} \cmidrule(lr){8-10}
    ~&\textbf{CSR}&\textbf{ISR}&\textbf{PSR}&\textbf{CSR}&\textbf{ISR}&\textbf{PSR}&\textbf{CSR}&\textbf{ISR}&\textbf{PSR} \\
\midrule
%o1-mini$^{\dag}$    &0.932&0.804&0.840	     &0.801&0.400&0.564	     &0.867&0.602&0.702\\
o1-preview$^{\dag}$    &0.926&0.806&0.844	     &0.814&0.462&0.592	     &0.870&0.634&0.718\\
% gemini-thinking$^{\dag}$    &\textbf{0.}&\textbf{0.}&\textbf{0.}	     &\textbf{0.}&\textbf{0.}&\textbf{0.}	     &\textbf{0.}&\textbf{0.}&\textbf{0.}\\
DeepSeek-V3$^{\dag}$    & \uwave{0.948} & \uwave{0.836} & \uwave{0.864}     & \uwave{0.831} & \uwave{0.460} & \uwave{0.616}     & \uwave{0.890} & \uwave{0.648} & \uwave{0.740} \\
DeepSeek-R1$^{\dag}$    & \textbf{0.960} & \textbf{0.874} & \textbf{0.894}     & \textbf{0.856} & \textbf{0.524} & \textbf{0.672}     & \textbf{0.908} & \textbf{0.699} & \textbf{0.783} \\
GPT-4o$^{\dag}$    & \uline{0.956} & \uline{0.868} & \uline{0.888}     & \uline{0.816} & \uline{0.438} & \uline{0.582}     & \uline{0.886} & \uline{0.653} & \uline{0.735} \\

GPT-4-Turbo-20240409$^{\dag}$   &0.924&0.792&0.826	     &0.783&0.370&0.518	     &0.853&0.581&0.672\\
GPT-4-0125-Preview$^{\dag}$     &0.923&0.790&0.826	     &0.763&0.310&0.468	     &0.843&0.550&0.647\\
%GPT-3.5-Turbo-1106$^{\dag}$     &0.797&0.520&0.602	     &0.631&0.176&0.326	     &0.714&0.348&0.464\\
Claude-3.5-Sonnet$^{\dag}$      & 0.943 & 0.844 & 0.882     & 0.799 & 0.408 & 0.564     & 0.871 & 0.626 & 0.723 \\
GLM-4-0520$^{\dag}$             & 0.939 & 0.820 & 0.852     & 0.785 & 0.372 & 0.536     & 0.862 & 0.596 & 0.694 \\

% DeepSeek-V2-0628$^{\dag}$       &\uline{0.946}&0.830&0.868	     &0.786&0.350&0.524	     &0.866&0.590&0.696\\
ERNIE-4-Turbo-0628$^{\dag}$     &0.930&0.790&0.848	     &0.772&0.332&0.532	     &0.851&0.561&0.690	\\
Yi-Large$^{\dag}$               &0.900&0.730&0.786	     &0.744&0.292&0.460	     &0.822&0.511&0.623\\
% abab6.5-chat$^{\dag}$           &0.894&0.696&0.766	     &0.736&0.260&0.452	     &0.815&0.478&0.609\\
MoonShot-V1-8k$^{\dag}$         &0.919&0.764&0.812	     &0.758&0.308&0.464	     &0.838&0.536&0.638\\
\midrule
Llama-3-8B-Instruct$^{\ast}$     &0.656&0.300&0.356	     &0.562&0.122&0.238	     &0.609&0.211&0.297\\
Llama-3-70B-Instruct$^{\ast}$    &0.750&0.422&0.498	     &0.642&0.178&0.330	     &0.696&0.300&0.414\\
\midrule
DeepSeek-V2-Lite-Chat       &0.733&0.382&0.448	&0.597&0.148&0.262	&0.665&0.265&0.355\\
% I-1.5-34B-Chat             &0.881&0.672&0.740	&0.745&0.302&0.474	&0.813&0.487&0.607\\
Qwen1.5-110B-Chat       &0.905&0.724&0.792	&0.730&0.276&0.438	&0.818&0.500&0.615\\
Qwen2-72B-Instruct      & 0.944 & 0.836 & 0.880     & 0.791 & 0.342 & 0.530     & 0.867 & 0.589 & 0.705 \\

\bottomrule[0.1em]
\end{tabularx}
\caption{\label{Experiments Overall Result}
 The evaluation results of LLMs on CFBench and its splits. Notably, $^{\ast}$ stands for the model supporting mainstream languages excluding Chinese, and $^{\dag}$ represents calling through the API. The \textbf{bold}, \uline{underlined}, and \uwave{tilde} denote the first, second, and third rankings, respectively. 
}
\end{table*}
% %实验结果

\subsection{Dataset Statistics}
\subsubsection{Overall Statistics}
Table \ref{CFBench overall Statistic} provides a statistical overview of CFBench, highlighting substantial differences between the two sets. The Hard Set has more detailed instructions, a greater variety, and a higher number of constraints, indicating higher complexity compared to the Easy Set. The table also shows the diversity and balanced distribution of primary constraint types within CFBench, outperforming other benchmarks. See the Appendix for division details.

\subsubsection{Tasks and Domains Distribution}
CFBench covers 20 major real-life domains and includes over 200 common scenarios and 50+ NLP tasks. Fig. \ref{Distribution of Domain and Scenario}(a) illustrates the classification of NLP tasks, including four major types: classification, generation, sequence labeling, and sentence relation, along with their corresponding specific tasks. Fig. \ref{Distribution of Domain and Scenario}(b) shows the real-life scenario-specific domain distribution, where Employment is the most prevalent category, and the other domains are relatively balanced. Our objective is to balance the real distribution with an equitable distribution. Overall, Fig. \ref{Distribution of Domain and Scenario} illustrates that CFBench has evolved into a comprehensive and well-balanced benchmark.

\subsubsection{Comparison with Other Benchmarks}
As shown in Tab. \ref{Comparison}, we thoroughly compare our benchmark with various relevant ones. In terms of size, our benchmark contains approximately twice the number of samples as others. FollowBench~\cite{jiang2023followbench} increases difficulty by adding the number of constraints, but focuses on the incremental increase of a single constraint type. ComplexBench~\cite{wen2024benchmarking} places more emphasis on the combination relationships between different constraint types, but only designs four types. IFEval~\cite{zhou2023instruction} focuses on constraints that can be verified, but lacks generalization. Compared to others, CFBench provides comprehensive scenario coverage, diverse systematic constraints, numerous high-quality samples, and multidimensional evaluation. For details, see Appendix Tab.~\ref{Benchmark Comparison}, including Case and Features.

%Notably, FollowBench, despite having 820 samples, originally had only one-fifth of this number, with each sample expanded by adding five constraints. From the constraint perspective, CFBench is the only benchmark with systematized constraints, featuring the highest number of primary constraint types and ensuring a balanced consideration of these types. Regarding evaluation, InFoBench also sets response criteria for each sample, while CFBench distinguishes itself by incorporating user demand prioritization and emphasizing objectivity

%,including IFEval\cite{zhou2023instruction}, CELLO~\cite{he2024can}, FollowBench~\cite{jiang2023followbench}, and InfoBench~\cite{qin2024infobench}

\subsection{Evaluation Protocol}

\subsubsection{Evaluation Criteria}
We breaking down instructions into multiple simple, independent checkpoints to ensure evaluation accuracy, inspiration was drawn from DRFR~\cite{qin2024infobench}. %This approach addresses the challenge of evaluating entire responses, especially for complex instructions with multiple constraints. 
Unlike DRFR, our method emphasizes defining ideal response characteristics and critical evaluation points. The previous sections detailed the checklist generation process, a key part of our evaluation criteria. Furthermore, we employ GPT-4o, as the evaluation model. By repeatedly feeding it the instruction, test model response, and checklist with a carefully tuned prompt, we ensure that the judged response fully meets the judgement format check. This iterative process aims to maximize confidence in our evaluation. The specific evaluation prompt is in the Appendix. 

\subsubsection{Evaluation Metrics}
Aligned with different perspectives, we define the Constraint Satisfaction Rate (CSR), Instruction Satisfaction Rate (ISR) as follows:

\begin{equation}
\mathrm{CSR}=\frac1{m}\sum_{i=1}^m(\frac1{n_i}\sum_{j=1}^{n_i}s_i^j)
\end{equation}

\begin{equation}
\mathrm{ISR}=\frac1{m}\sum_{i=1}^ms_i 
\end{equation}

where $s_i^j$ = 1 if the $j$-th constraint of $i$-th instruction is satisfied and $s_i^j$ = 0 otherwise. $s_i$ = 1 indicates that all constraints in the $i$-th instruction are satisfied and $s_i$ = 0 otherwise. The requirements Priority Satisfaction Rate (PSR) is defined as follows:
\begin{equation}
\mathrm{PSR}=\frac1{m}\sum_{i=1}^m(PSR_i)
\end{equation}
Let the average score for secondary requirements be $A$. When all primary requirements are met, $score = 0.5 + 0.5 \times A$. If $score > 0.8$, then $PSR_i = 1$; otherwise, $PSR_i = 0$, 
especially when any primary requirement is not met. The threshold of $0.8$ is based on user feedback, reflecting tolerance for LLMs adhering to constraints. Overall, CSR, ISR, and PSR reflect different levels of user perception from multiple perspectives, including constraints, instructions, and requirement priorities.

% % 指令跟随相关benchmark的详细对比
\begin{table}
\centering
\resizebox{\linewidth}{!}{ % 自动调整表格宽度
\footnotesize % 调整字体大小
\begin{tabularx}{\linewidth}{p{1.5cm}ccccc} % 缩小第一列宽度
\toprule[0.1em]
\multirow{2}*{\textbf{Benchmarks}} & \multicolumn{3}{c}{\textbf{Data Quality}} & \multicolumn{2}{c}{\textbf{Evaluation}} \\
\cmidrule(lr){2-4} \cmidrule(lr){5-6}
~ &\textbf{Num.} &\textbf{Type.} &\textbf{Syst.} &\textbf{Prio.} &\textbf{Meth.} \\
\midrule
IFEval  &541 &\hspace{3pt}4$^{\ast}$  &\XSolidBrush &\XSolidBrush &\includegraphics[width=0.015\textwidth]{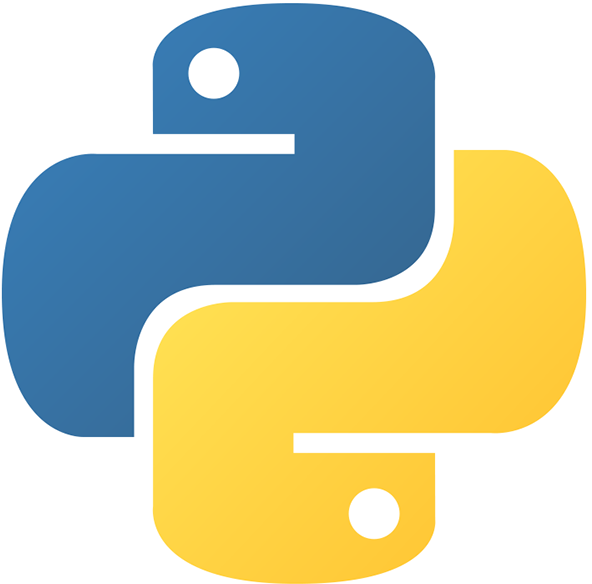}   \\
CELLO  &523 &4  &\XSolidBrush &\XSolidBrush &\includegraphics[width=0.015\textwidth]{cfbench_fig/tubiao-py.png}   \\
FollowBench  &820 &5  &\XSolidBrush &\XSolidBrush &\includegraphics[width=0.015\textwidth]{cfbench_fig/tubiao-py.png}\includegraphics[width=0.015\textwidth]{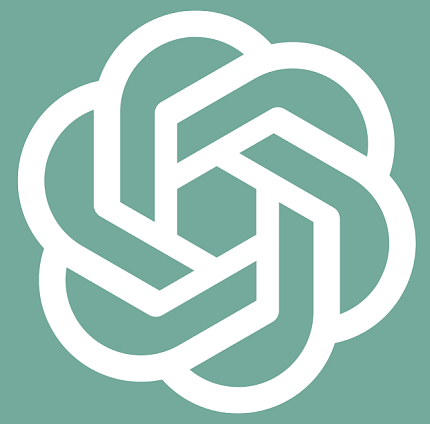}    \\
InFoBench  &500 &5 &\XSolidBrush &\XSolidBrush &\includegraphics[width=0.017\textwidth]{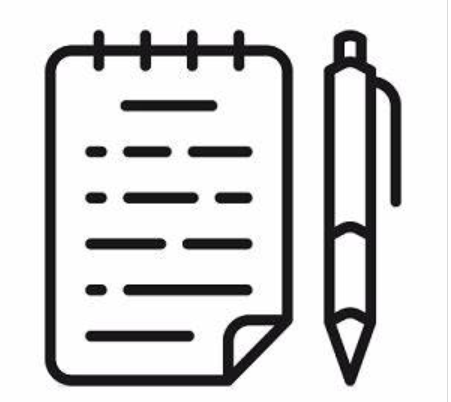}\includegraphics[width=0.015\textwidth]{cfbench_fig/tubiaoGPT4.png}    \\
FoFoBench  &494 &1 &\XSolidBrush &\XSolidBrush &\includegraphics[width=0.015\textwidth]{cfbench_fig/tubiaoGPT4.png}   \\
ComplexBench  &1150 &4 &\CheckmarkBold &\XSolidBrush &\includegraphics[width=0.015\textwidth]{cfbench_fig/tubiaoGPT4.png}   \\
\rowcolor[gray]{0.85} CFBench  &1000 &10-25 &\CheckmarkBold &\CheckmarkBold &\includegraphics[width=0.0170\textwidth]{cfbench_fig/tubiao-crit.jpg}\includegraphics[width=0.015\textwidth]{cfbench_fig/tubiaoGPT4.png}   \\
\bottomrule[0.1em]
\end{tabularx}
}
\caption{\label{Comparison} Detailed Comparison of Relevant Benchmarks. $^{\ast}$ represents our constraint system. 'Num.', 'Type.', 'Syst.', 'Prio.', and 'Meth.' denote the number of samples, primary constraint types, presence of a constraint system, requirement prioritization, and evaluation method.}
\vspace{-10pt} % 调整这个值来减少空白
\end{table}

% % abbreviations respectively

\subsection{Data Quality}
\subsubsection{Quality Evolution}
To enhance the quality of CFBench, we invested considerable effort and financial resources. First, in the instruction generation phase, we utilized multiple advanced LLMs, such as GPT-4 and Claude, to generate diverse instructions and responses for annotator candidates. Second, we implemented a stringent manual annotation process, including annotator training, cross-validation, batch validation, expert team involvement, and iterative refinement of instruction constraints and response quality. We also ensured the objectivity, evaluability, and prioritization of checkpoints. Additionally, we balanced the data for constraint types, scenarios, and NLP task distribution. Detailed information can be found in the Appendix. 

\subsubsection{Quality Evaluation}
To investigate CFBench's quality, we randomly selected 100 samples for assessment. Three professional data inspectors evaluated them, resulting in high-quality rates of 94\% for instructions, 94\% for gold answers, and 93\% for checklists (see Appendix Table \ref{High-Quality Rate}). Additionally, three experts rated Qwen2-7B-Instruct outputs on a 0-1 scale. The kappa coefficient between GPT-4o PSR and expert evaluations was 0.77, highlighting the effectiveness of the PSR evaluation method and metrics, even for smaller models. Further details are in Appendix Table \ref{kappa coefficient}.
%In conclusion, CFBench is a high-quality benchmark, both in terms of data and evaluation.   

\begin{figure}[t]
\centering
\includegraphics[width=0.9\columnwidth]{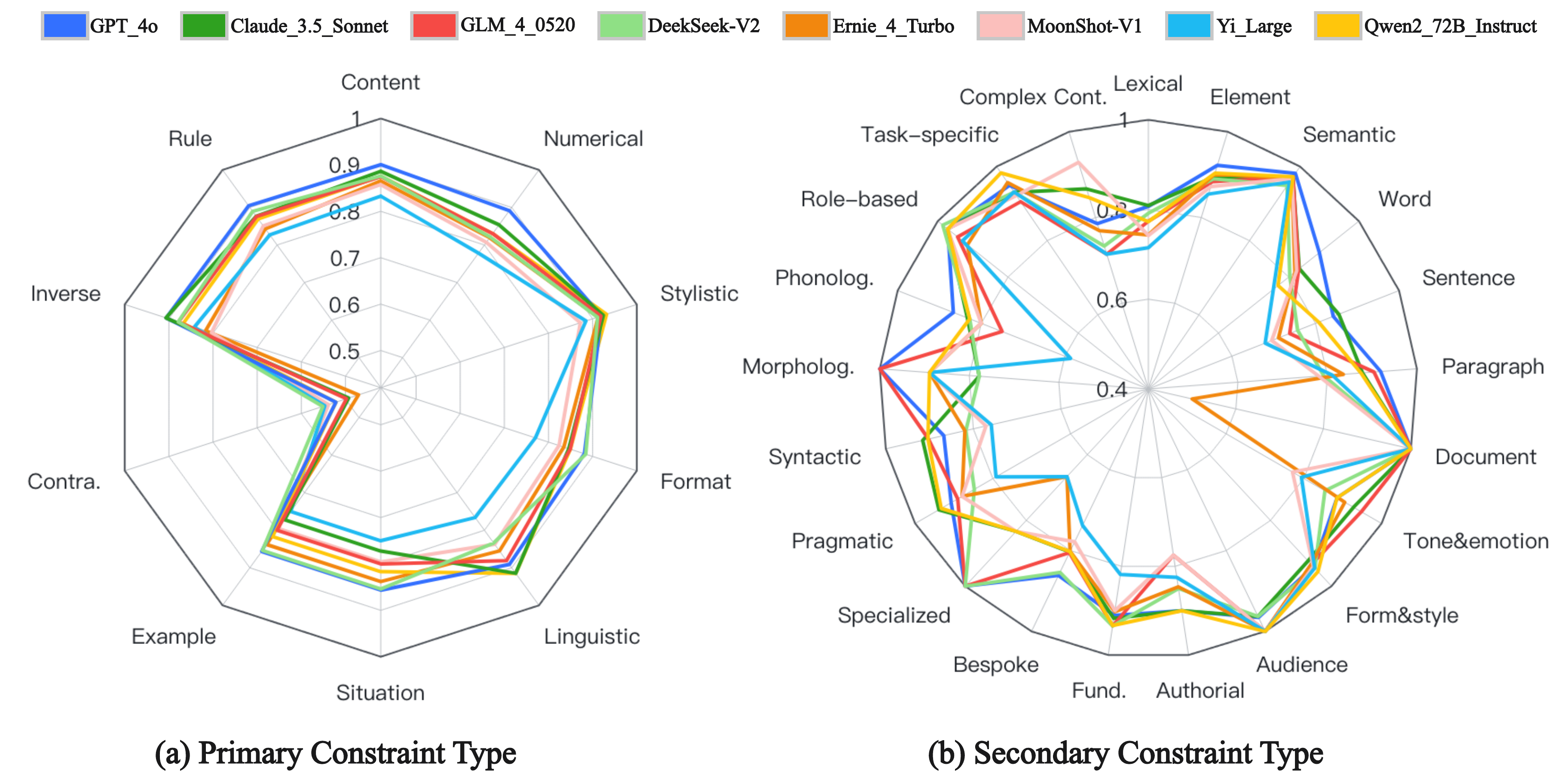} % Reduce the figure size so that it is slightly narrower than the column. Don't use precise values for figure width.This setup will avoid overfull boxes.
\caption{Different mainstream models' results under primary and secondary constraint categories.}
\label{constraints type result}
\end{figure}

\begin{figure}[t]
\centering
\includegraphics[width=0.9\columnwidth]{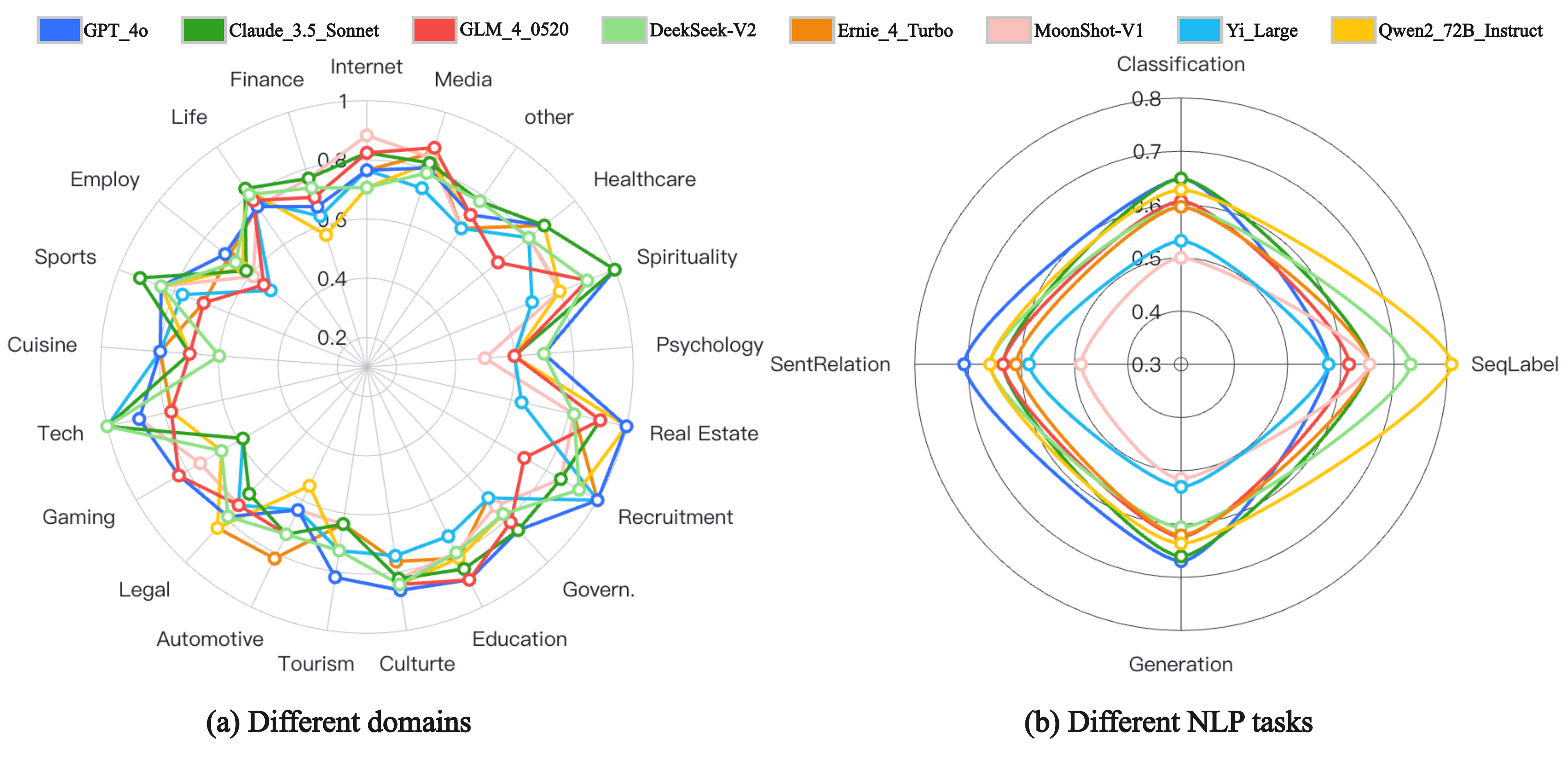} % Reduce the figure size so that it is slightly narrower than the column. Don't use precise values for figure width.This setup will avoid overfull boxes.
\caption{Different mainstream models' PSR results in real-world domains and NLP task types.}
\label{domain result} 
\end{figure}

\section{Experiment}
\subsection{Evaluation Settings}
We evaluated 50+ top-performing models from previous benchmarks \cite{hendrycks2020measuring, cobbe2021training}, considering factors like model size, Chinese language support, access via API or weights, and fine-tuning with instruction data. During inference, we set the maximum generation length to 2048 and used default values for other parameters. For evaluation, we used GPT-4o as the judge model with a temperature of 0 for deterministic outputs.

% % 表格:不同方向benchmark对比

% 图:讨论-影响约束指令跟随效果的因素
\begin{figure*}[t]
\centering
\includegraphics[width=0.9\textwidth]{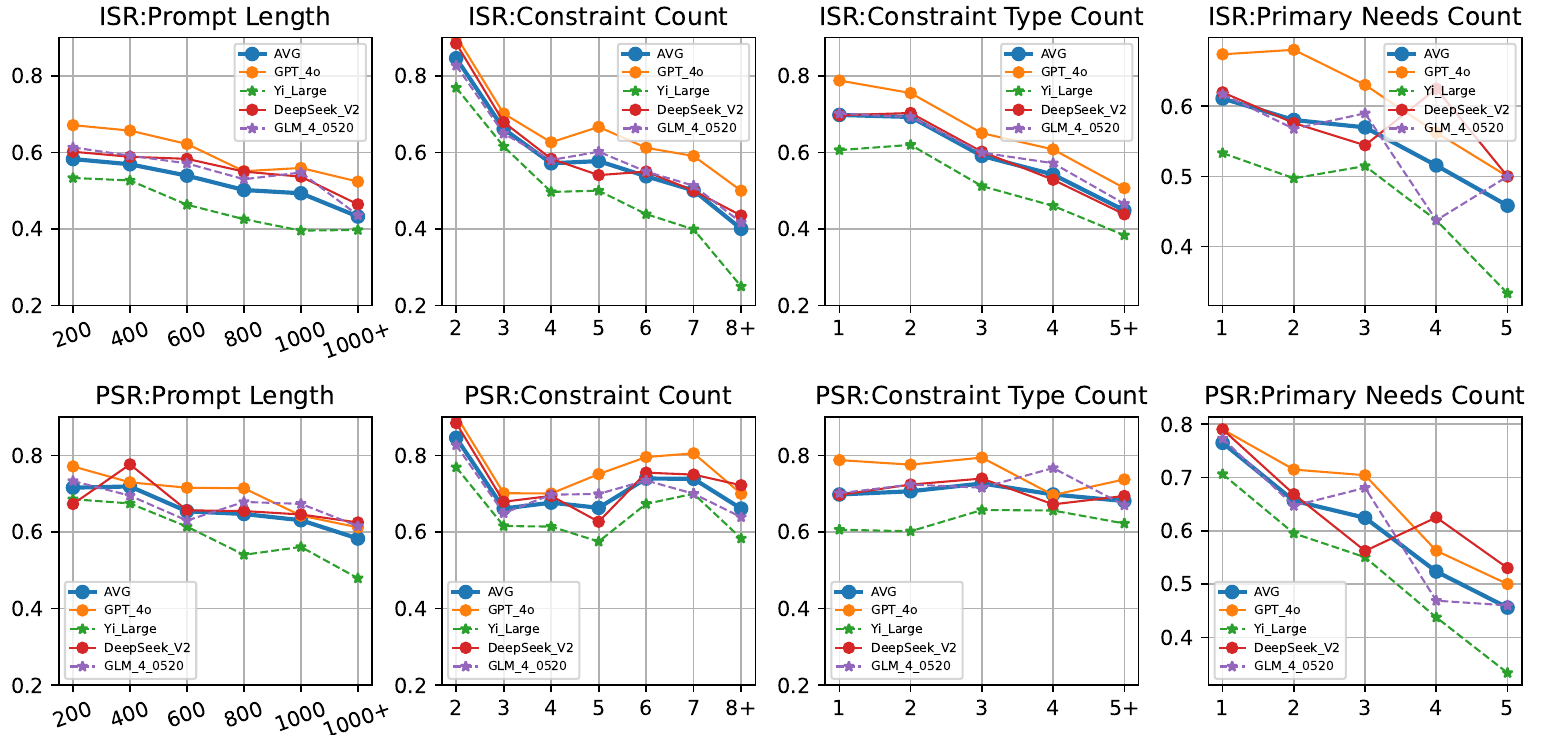} % Reduce the figure size so that it is slightly narrower than the column.
\caption{Factors Influencing Constraints-Following Performance}
\label{Factors}
\end{figure*}
% 图:讨论-影响约束指令跟随效果的因素

\subsection{Overall Result}

Tab. \ref{Experiments Overall Result} presents CFBench evaluation results for leading models. DeepSeek-R1~\cite{deepseekai2024deepseekv3technicalreport} leads overall, followed by GPT-4o and DeepSeek-V3~\cite{deepseekai2025deepseekr1incentivizingreasoningcapability} in third. Claude-3.5-Sonnet~\cite{anthropic2024claude}  and Qwen2-72B-Instruct~\cite{yang2024qwen2} performed well, though DeepSeek-V3 and Qwen2-72B-Instruct showed slight drops in the Full Set. The highest PSR in the Hard Set was 0.582, indicating room for improvement.

While CSR favors weaker models, ISR and PSR highlight differences in stronger models. API-accessed models like GPT-4o outperformed most open-source models, though DeepSeek-V3 and Qwen2-72B-Instruct performed well among open-source models.

% % 表格:不同方向benchmark对比
\begin{table}
\centering
\small % 调整字体大小
\begin{tabularx}{\linewidth}{p{2.8cm}ccc}
\toprule[0.1em]
\textbf{Model} &\textbf{MMLU} &\textbf{GSM8K} &\textbf{CFBench} \\
\midrule
GPT-4o              &88.7&90.5&0.698\\
Claude-3.5-Sonnet	&88.7&96.4&0.691\\
Qwen2-72B-Instruct	&82.3&91.1&0.672   \\
% GLM-4	&81.5&87.6&0.665 \\
DeepSeek-V2	&78.5&79.2&0.665\\
Qwen1.5-110B-Chat	&80.4&-&0.584\\
Qwen1.5-72B-Chat	&77.5&79.5&0.577\\
\bottomrule[0.1em]
\end{tabularx}
\caption{\label{Performance Comparison on Benchmarks}Performance Comparison on Benchmarks}
\end{table}

\subsection{Constraints-categorized Performance}

To assess performance across different constraint types, we calculated satisfaction scores for the top 8 LLMs (see Fig. \ref{constraints type result}). Many models struggled with contradictory constraints, highlighting their limitations. GPT-4o excelled across various constraints, while other models alternated in leading different types. For secondary constraints, all models performed poorly in lexical, word, and sentence count constraints but did better in document count and audience style constraints. No single model consistently led across most constraint types. In summary, even the most advanced LLMs have significant room for improvement, with each model showing specific weaknesses, providing valuable insights for future iterations.

\subsection{Domain and Task-categorized Performance}
As depicted in Fig. \ref{domain result}, we evaluate performance across 21 domains and 4 major NLP task types, each with 500 examples from the two main sources of CFBench. For domain performance, employment and psychology require significant attention, while technology and recruitment are strengths for most models. For NLP tasks, GPT-4o excels in sentence relationship tasks, while Qwen2-72B-Instruct is strong in sequence labeling, likely due to its optimization for Chinese. In general, models exhibit different rankings across domains and tasks, indicating no clear absolute leader. Comprehensive improvements are needed for better constraint follow across multiple domains and tasks.

\section{Discussions}

\subsection{Comparisons between Capabilities}
Table \ref{Performance Comparison on Benchmarks} presents a comprehensive comparison of CFBench's PSR with two prominent LLM evaluation benchmarks: MMLU~\cite{hendrycks2020measuring} and GSM8K~\cite{cobbe2021training}. MMLU focuses on knowledge proficiency, while GSM8K emphasizes mathematical ability. GPT-4o ranks first on CFBench but significantly lags behind, ranking third on GSM8K. 
Qwen2-72B-Instruct performs worse than DeepSeek-V2 on CFBench but outperforms it on MMLU. Notably, the rankings of LLMs on CFBench do not entirely correspond with those on the other two benchmarks, indicating that CFBench provides a novel perspective for LLMs evaluation. 

\subsection{Factors influencing constraints-following}
Previously, we identified a significant gap in LLM constraints following performance, prompting us to further explore the influencing factors. We analyzed the impact of prompt length, number of constraints, constraint types, and primary requirements on evaluation results across five top-performing models and their average values. As shown in Fig. \ref{Factors}, all four factors are positively correlated with the ISR metric, with the number of constraints having the most significant effect. For PSR, the number of constraints and constraint types do not show a completely positive correlation, while the number of primary requirements has a greater influence. Users are more affected by unmet constraints when there are fewer, but become more tolerant of unmet non-primary constraints when there are many.   

\subsection{How to improve constraint-following ability?}
In Appendix Tab. \ref{Complete Experiments Result}, we investigated methods to potentially enhance constraint following. Firstly, Supervised Fine-Tuning (SFT) significantly improves performance, with nearly all models that undergo instruction fine-tuning exhibiting substantial improvements in effectiveness, as demonstrated by the Qwen series. Secondly, model size is also an important factor, as evidenced by Qwen2-72B-Instruct showing a 40\% relative PSR improvement over Qwen2-7B-Instruct.  Additionally, replicating Conifer's models ~\cite{sun2024conifer} reveals that fine-tuning with complex constraint instructions further enhances performance, and recent work has also been directed towards this approach ~\cite{he2024complex}. Further exploration is intended to be pursued in future work.  

\section{Conclusion}
This study comprehensively examines the constraints-following capabilities of LLMs. CFBench, a comprehensive benchmark, was introduced with 1000 manually annotated samples covering more than 200 real-world scenarios and over 50 NLP tasks, encompassing a wide range of systematically defined constraint types. Each sample in CFBench includes detailed evaluation criteria, providing metrics that accurately reflect model performance and real user needs across multiple dimensions. Extensive experiments on CFBench revealed significant limitations and challenges that advanced LLMs face in following constraint instructions. Key factors and potential strategies to improve constraint following were also analyzed, and numerous insightful findings can provide valuable guidance for the optimization of LLMs' performance. In conclusion, CFBench offers a novel perspective for evaluating LLM capabilities, providing new directions for performance assessment and improvement.   

\section{Limitations}
\subsection{Experimental Setup}
This study primarily focuses on models with strong Chinese language capabilities, lacking a comprehensive survey of a broader range of English models. Additionally, while we conducted preliminary analyses on the differences in instruction-following abilities between Chinese and English, a more in-depth comparative study is absent.

\subsection{Limited Exploration of Reasoning Models}
Currently, deep reasoning models like R1 continue to achieve commendable results. However, there is a lack of in-depth research into these models, particularly concerning the factors that enhance their instruction-following abilities.

\subsection{Evaluation Model Bias}
The evaluation of models predominantly relies on GPT-4o as the judge model. Future research could explore the impact of different evaluation models on assessing the performance of other models.

\section{Ethics Statement}
This research adheres to the ethical guidelines set forth by the Association for Computational Linguistics (ACL). We have ensured that all data collection and experimental designs comply with privacy protection and informed consent principles, fully respecting and safeguarding the rights of all participants. Furthermore, we have evaluated the potential societal impacts of our research findings, ensuring that their application does not result in adverse effects on society.
\newpage
\newpage
% \bibliography{aaai25}
\bibliography{custom}
\appendix

\section{Appendix}
\subsection{Constraint System Construction}
\begin{figure}[t]
\centering
\includegraphics[width=0.9\columnwidth]{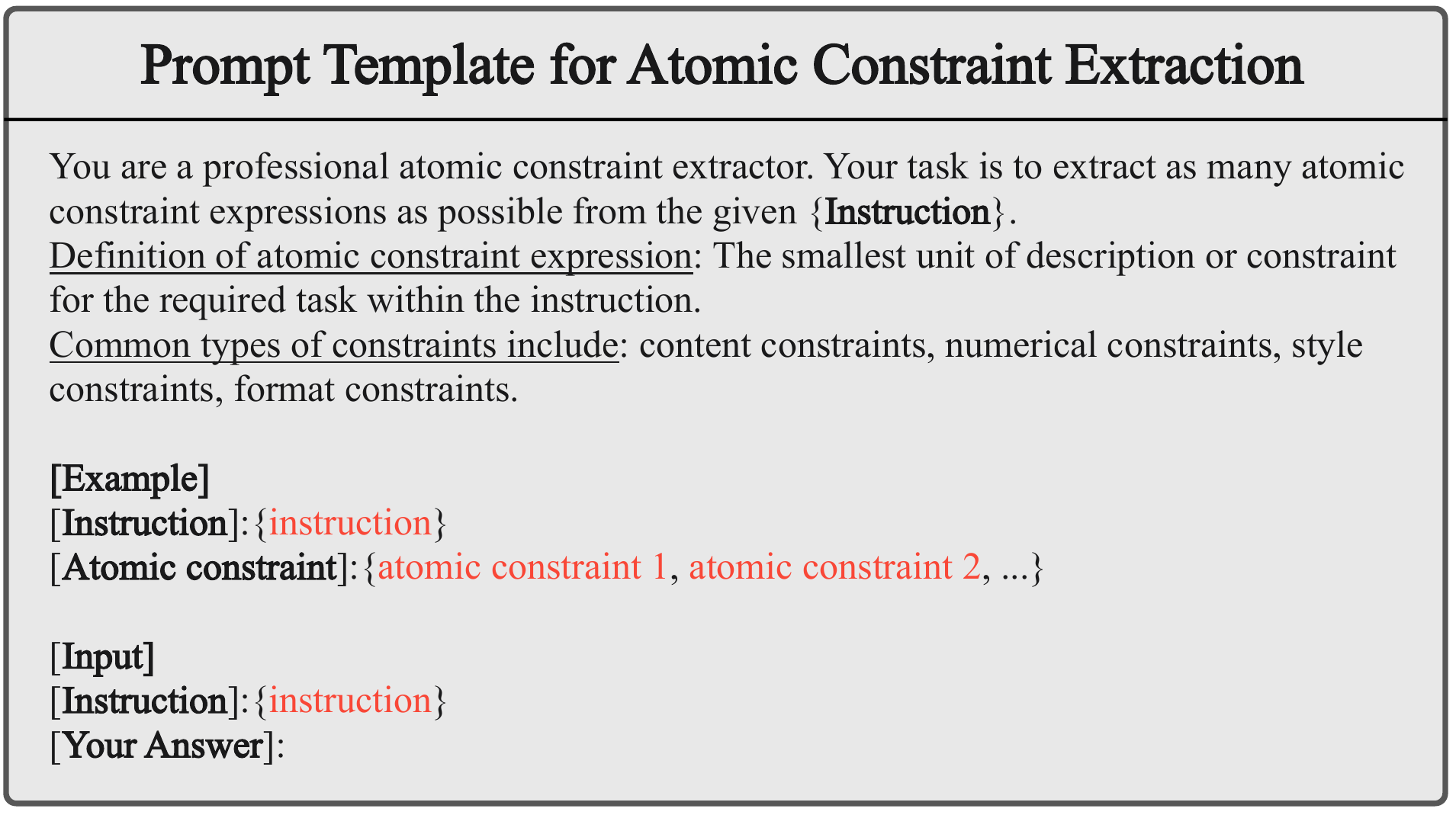}
\caption{Prompt Template Atomic Constraint Extract}
\label{atomic constraint extraction}
\end{figure}

The construction of the constraint system commenced with the aggregation of data from diverse real-world scenarios and NLP tasks. This encompassed 800,000 query logs from LLM websites over the preceding six months, alongside over 300,000 data points from various NLP tasks. Instructions that were excessively long or short were filtered out, and a vector clustering deduplication algorithm was employed. This meticulous process culminated in a refined dataset comprising approximately 30,000 instructions. Subsequently, GPT was utilized to extract constraint atoms from these instructions, thereby ensuring the comprehensiveness of the constraint system. The prompt employed for GPT-4 extraction, as illustrated in Fig. \ref{atomic constraint extraction}, resulted in the identification of approximately 5,000 unique atomic constraint instructions. Three seasoned experts meticulously refined these into 1,000 significant atomic constraints. By integrating statistical analysis, classification, and linguistic principles, a hierarchical constraint system was developed using Top-Down Organization and Bottom-Up Synthesis methodologies. This system comprises 10 primary categories and 25 secondary categories. The system comprehensively categorizes all types of constraints, ensuring that nearly all specific constraints can be systematically classified within its framework. Detailed information regarding the constraint system is presented in Tab. \ref{constraint system}.  

\subsection{Dataset Construction}
\begin{figure}[t]
\centering
\includegraphics[width=0.9\columnwidth]{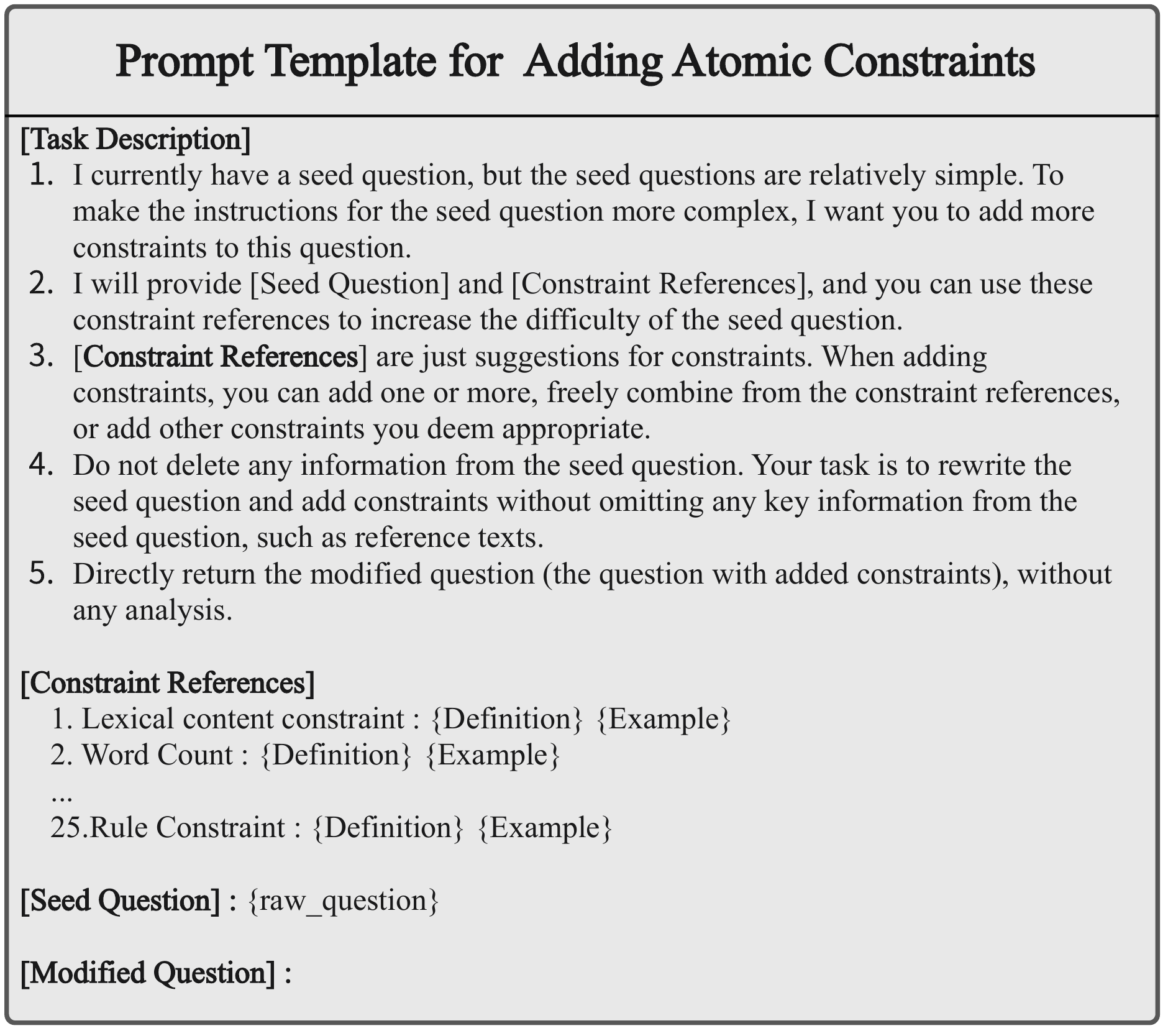} 
\caption{Prompt Template for Adding Atomic constraint}
\label{add atomic constraint}
\end{figure}

\begin{figure*}[t]
\centering
\includegraphics[width=0.9\textwidth]{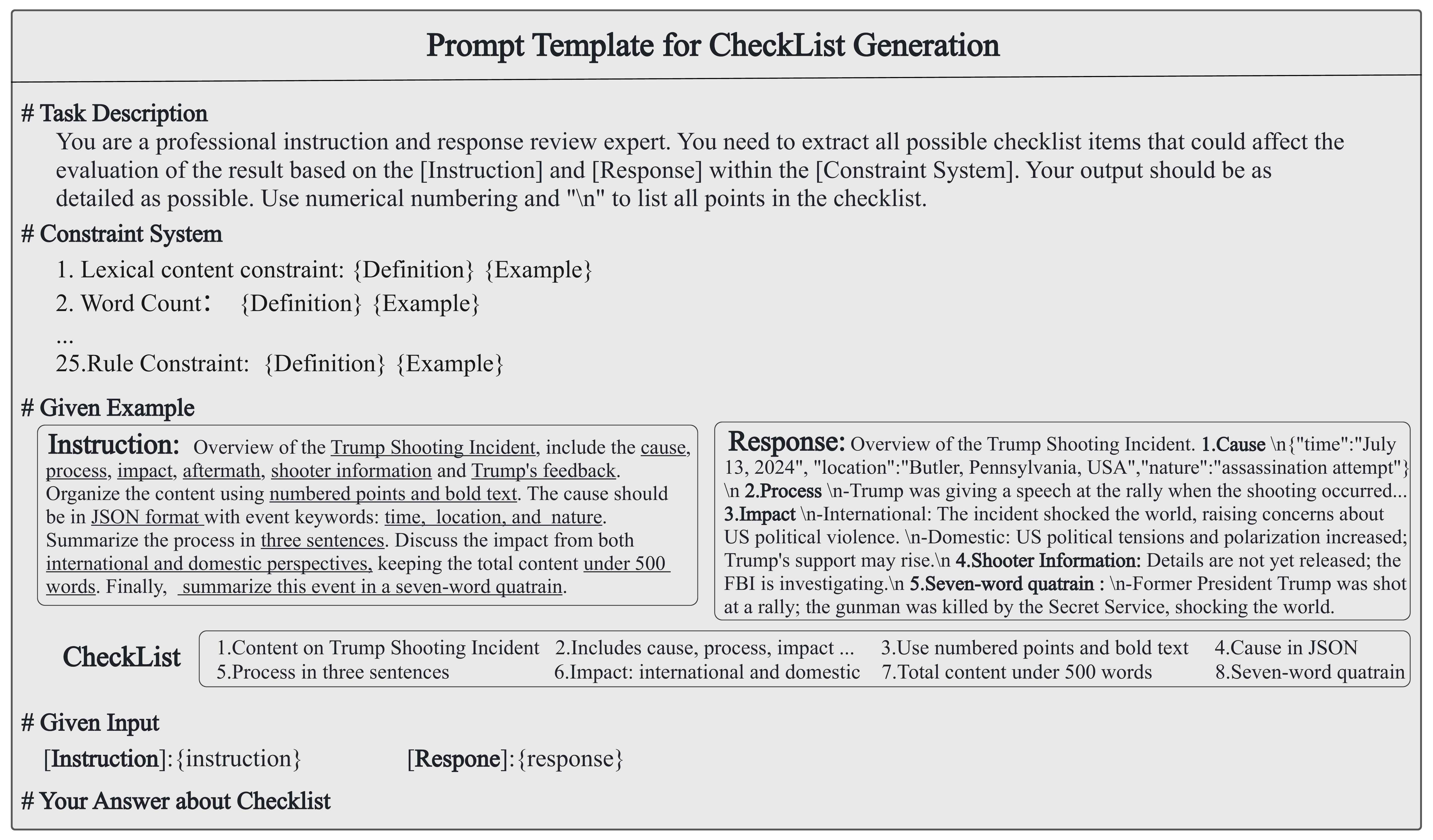} % Reduce the figure size so that it is slightly narrower than the column.
\caption{Prompt Template for CheckList Generation}
\label{checlist generation prompt}
\end{figure*}

We adopted an innovative Human-Machine Collaborative Iterative Construction approach to ensure the highest quality of data. This method involved leveraging advanced LLMs to augment original instructions with additional constraints and generate corresponding responses. These responses were meticulously reviewed for constraint validity, followed by the creation of detailed checklists for each example. Multiple experts participated in this iterative process, continuously refining the outputs by addressing issues encountered by the LLMs and regenerating or manually correcting any substandard samples. The prompts used for GPT to enhance constraints and generate checklists are illustrated in Fig. \ref{add atomic constraint} and Fig. \ref{checlist generation prompt}. Due to the limited attention given to real-life scenarios, we have meticulously organized and covered 20 domains and over 200 scenarios in our CFBench system, as detailed in Tab. \ref{Domain and Scenarios List}. In the end, we gathered 1,000 high-quality data points: 500 from real-world scenarios and 500 from different NLP tasks.  Specifically, we implemented the following steps to enhance data quality for manual annotations.

\subsubsection{Annotator Training}
We sourced annotation contractors from the public and selected 21 candidates for training by seasoned data scientists. After a one-week training program, the annotators engaged in multiple rounds of trial annotations, which were then assessed by data experts. From these assessments, the 9 annotators demonstrating the highest accuracy were selected for this dataset.
\subsubsection{Cross-Validation}
To reduce the likelihood of missed and incorrect annotations, we implemented an inter-annotator validation process. Three annotators independently reviewed the labeled instructions, responses, and evaluation criteria, achieving a notable agreement rate of 94\%. Any discrepancies that emerged were resolved through expert adjudication, ensuring both consistency and accuracy.
\subsubsection{Batch Validation}
Due to the substantial size of the dataset, it was systematically divided for processing. Following a phased improvement approach, the initial batch sizes were set at 50, 100, and 200, gradually increasing to 400 for later batches. After the annotation process, 50\% of the dataset was randomly selected for contractor review, while 20\% of the dataset was examined by experts.
\subsubsection{Data Split}
We used a voting mechanism involving experts and ten models, including GPT-4o and Claude-3.5-Sonnet, to partition 10,000 CFBench entries into 'easy' and 'hard' categories. The 'hard' category includes entries where multiple models struggle with PSR performance and are also challenging for humans, as verified by experts. 
%This partitioning maps model performance across difficulty levels: 'hard' challenges advanced models, while 'easy' targets average models.

\subsection{Evaluation Method and Metric}
\begin{figure}[t]
\centering
\includegraphics[width=0.9\columnwidth]{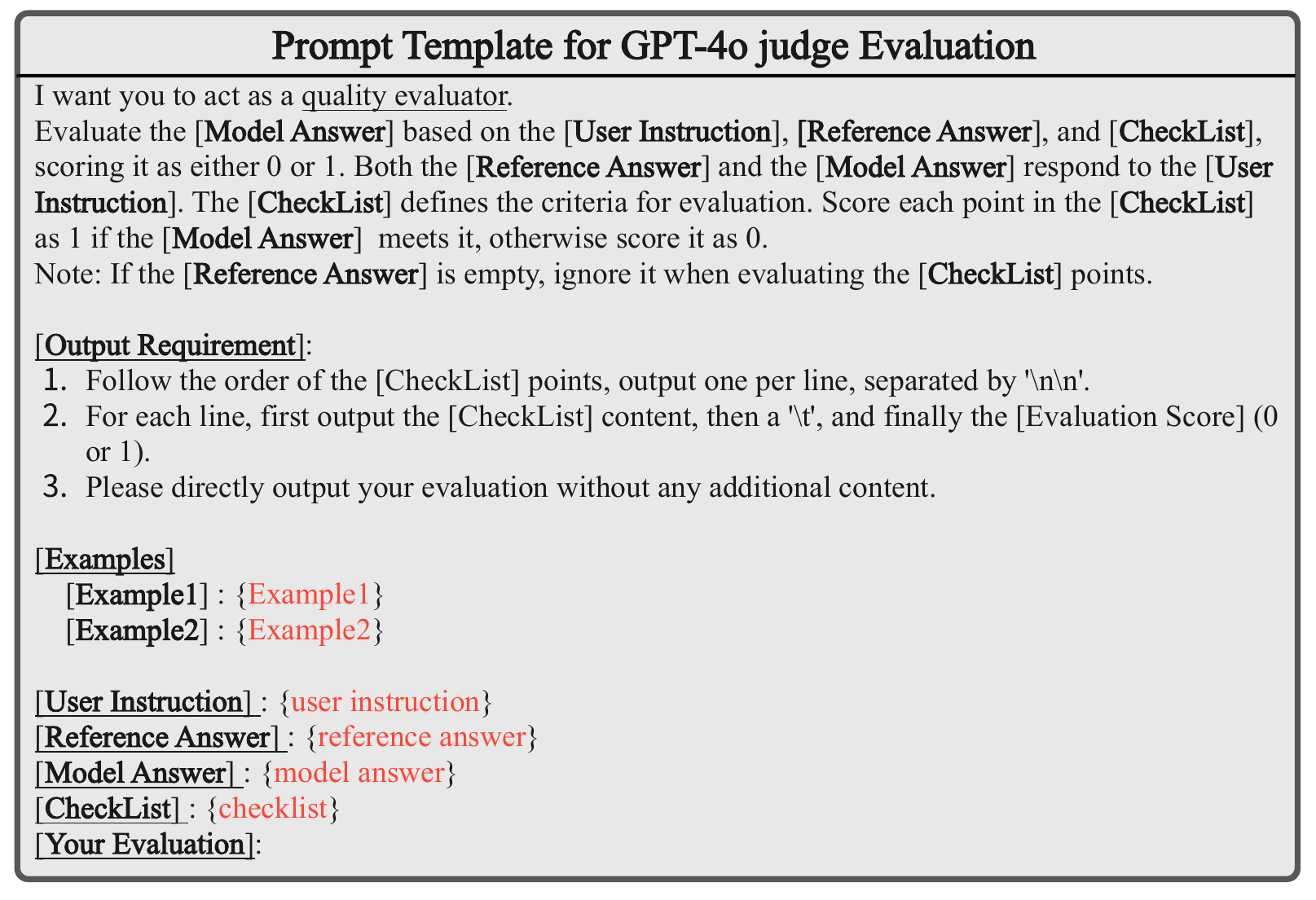}
\caption{Prompt Template for Evaluation}
\label{Prompt Template for Evaluation}
\end{figure}

The state-of-the-art GPT-4o model was employed as the judge to perform binary scoring (0 or 1) for each checkpoint in the checklist. The specific evaluation prompt is illustrated in Fig. \ref{Prompt Template for Evaluation}. The Requirement Priority-Satisfaction Ratio (PSR) was proposed as an evaluation metric that simultaneously considers the prioritization of user requirements and satisfaction levels. PSR is calculated by first ensuring that all primary requirements are met. Subsequently, the satisfaction score is determined by averaging the fulfillment of the remaining constraints to obtain $A$. The final satisfaction score is then calculated using the formula $0.5 + 0.5*A$. If the final score exceeds $0.8$, PSR is set to $1$. The threshold of $0.8$ was established based on the average satisfaction levels derived from multiple users' feedback on the responses to the instructions. 

\subsection{Quality Assessment}
We employed multiple methods to validate the quality of the benchmark on a randomly selected set of 100 samples. First, we engaged three experts to independently evaluate the quality of each sample's instruction, response, and criteria. The average quality rate determined by the three experts was consistently above 90\%, as detailed in Tab. \ref{High-Quality Rate}. To further validate the effectiveness of our proposed evaluation metric, PSR, we had the same three experts score the responses of Qwen2-7B-Instruct on these 100 cases using a 0-1 scale. Simultaneously, we utilized GPT-4o to directly score the responses, referred to as GPT-4o PSR. By calculating the kappa coefficient, we found a strong agreement between our proposed PSR evaluation metric and the human experts' assessments. The detailed results are presented in Tab. \ref{kappa coefficient}. Kappa coefficient scores are interpreted as follows: below 0.2 indicates slight agreement, 0.21 to 0.40 indicates fair agreement, 0.41 to 0.60 indicates moderate agreement, 0.61 to 0.80 indicates substantial agreement, and 0.81 to 1.00 indicates almost perfect agreement.

% CFBench数据质量 A.1
\begin{table}
\centering
\footnotesize % 更小的字体
\begin{tabularx}{\linewidth}{p{2cm}ccc}
\toprule[0.1em]
\textbf{Set} &\textbf{Instruction} &\textbf{Gold Ans} &\textbf{CheckList} \\
\midrule
Easy Set  &0.96 &0.94  &0.93   \\
Hard Set  &0.92 &0.95  &0.93   \\
All Set  &0.94 &0.94  &0.93    \\
\bottomrule[0.1em]
\end{tabularx}
\caption{\label{High-Quality Rate}The High-Quality Rate of 100 selected Samples}
\end{table}

% CFBench数据质量 A.1
\begin{table}
\centering
\footnotesize % 更小的字体

\begin{tabularx}{\linewidth}{p{2.9cm}ccc}
\toprule[0.1em]
\textbf{Set} &\textbf{Easy Set} &\textbf{Hard Set} &\textbf{Full Set} \\
\midrule
Avg.Expert &1 &1  &1   \\
GPT-4o DS  &0.58 &0.61  &0.60   \\
GPT-4o PSR  &0.76 &0.77  &0.77    \\
Qwen2-72B-Inst. PSR  &0.70 &0.73  &0.72    \\
\bottomrule[0.1em]
\end{tabularx}
\caption{\label{kappa coefficient} The kappa coefficient between expert evaluations and various assessment methods}
\end{table}

\subsection{Experimental Setup and Results}
\subsubsection{Experiment Setting}
We evaluated the most popular Large Language Models (LLMs), with the majority of these models being developed by companies based in China, primarily  to accommodate our CFBench's focus on the Chinese language. Among the 50 evaluated models, they can be categorized into two groups based on their access method: API-based and open-source weight-based models. It is worth noting that the Llama series models do not primarily support the Chinese language, which results in noticeably lower performance. Both conifer-base and conifer-test are based on the Mistral-7B foundational model. Llama-3-8B-Instruct-CN and Llama-3-70B-Instruct-CN respectively represent Llama-3-8B-Instruct-Chinese and Llama-3-70B-Instruct-Chinese, both of which have undergone Chinese SFT (Supervised Fine-Tuning). For the base models, we used a 3-shot approach to ensure a fair evaluation. The complete list of evaluated models can be found in Tab. \ref{Complete Experiments Result}.

\subsubsection{Explanation of Results}
GPT-4o and Claude3.5-Sonnet have demonstrated near-absolute leadership, achieving outstanding performance across various metrics and categories. Similarly, models such as GLM-4-0510, ERNIE-4-Bot-0613, ERNIE-4-Turbo-0628, DeepSeek-V2-0628, and Qwen2-72B-Instruct have also exhibited strong capabilities. Many models that support less mainstream Chinese languages performed significantly worse, which is unfair to them and only serves to illustrate their relative rankings. This also confirms that performance are highly correlated with language, especially within the scope of language constraints. From the perspective of open-source versus closed-source models, open-source models have generally achieved comprehensive success. However, Qwen2-72B-Instruct, as an open-source model, also demonstrated notable constraint-following capabilities. Regarding model size, within the Qwen series, performance metrics clearly improve with increasing model size. Additionally, models that have undergone Supervised Fine-Tuning (SFT) show significantly enhanced instruction-following capabilities. The complete evaluation results and rankings can be found in Tab. \ref{Complete Experiments Result}.

%实验结果：
\newcolumntype{C}{>{\centering\arraybackslash}X}
\begin{table*}
\centering
\footnotesize % 更小的字体

\begin{tabularx}{\textwidth}{p{3.8cm}CCCCCCCCCCC}
\toprule[0.1em]
    \multirow{2}*{\textbf{Models}} & \multicolumn{3}{c}{\textbf{Easy Set}} &\multicolumn{3}{c}{\textbf{Hard Set}} &\multicolumn{3}{c}{\textbf{Full Set}}\\    
    \cmidrule(lr){2-4} \cmidrule(lr){5-7} \cmidrule(lr){8-10}
    ~&\textbf{CSR}&\textbf{ISR}&\textbf{PSR}&\textbf{CSR}&\textbf{ISR}&\textbf{PSR}&\textbf{CSR}&\textbf{ISR}&\textbf{PSR} \\
\midrule
GPT-4o$^{\dag}$    &\textbf{0.956}&\textbf{0.868}&\textbf{0.888}	     &\textbf{0.816}&\textbf{0.438}&\textbf{0.582}	     &\textbf{0.886}&\textbf{0.653}&\textbf{0.735}\\
GPT-4-Turbo-20240409$^{\dag}$   &0.924&0.792&0.826	     &0.783&0.370&0.518	     &0.853&0.581&0.672\\
GPT-4-0125-Preview$^{\dag}$     &0.923&0.790&0.826	     &0.763&0.310&0.468	     &0.843&0.550&0.647\\
GPT-3.5-Turbo-1106$^{\dag}$     &0.797&0.520&0.602	     &0.631&0.176&0.326	     &0.714&0.348&0.464\\
Claude-3.5-Sonnet$^{\dag}$      &0.943&\uline{0.844}&\uline{0.882}	     &\uline{0.799}&\uline{0.408}&\uline{0.564}	     &\uline{0.871}&\uline{0.626}&\uline{0.723}\\
GLM-4-0520$^{\dag}$             &0.939&0.820&0.852	     &0.785&\uwave{0.372}&\uwave{0.536}	     &0.862&\uwave{0.596}&0.694\\
DeepSeek-V2-0628$^{\dag}$       &\uline{0.946}&0.830&0.868	     &0.786&0.350&0.524	     &0.866&0.590&0.696\\
ERNIE-4-Turbo-0628$^{\dag}$     &0.930&0.790&0.848	     &0.772&0.332&0.532	     &0.851&0.561&0.690	\\
ERNIE-4-Bot-0613$^{\dag}$       &0.929&0.792&0.832	     &0.779&0.338&0.518	     &0.854&0.565&0.675\\
ERNIE-3.5-0613$^{\dag}$         &0.901&0.720&0.772	     &0.758&0.302&0.482	     &0.830&0.511&0.627     \\
Yi-Large$^{\dag}$               &0.900&0.730&0.786	     &0.744&0.292&0.460	     &0.822&0.511&0.623\\
abab6.5-chat$^{\dag}$           &0.894&0.696&0.766	     &0.736&0.260&0.452	     &0.815&0.478&0.609\\
MoonShot-V1-8k$^{\dag}$         &0.919&0.764&0.812	     &0.758&0.308&0.464	     &0.838&0.536&0.638\\
\midrule
Vicuna-7B-V13$^{\ast}$           &0.563&0.206&0.262	     &0.468&0.100&0.168	     &0.516&0.153&0.215\\
Vicuna-33B-V13$^{\ast}$	         &0.621&0.270&0.352	     &0.527&0.110&0.196	     &0.574&0.190&0.274\\
Vicuna-13B-V13$^{\ast}$	         &0.605&0.248&0.302      &0.503&0.100&0.178	     &0.554&0.174&0.240\\
Llama-2-7B-Chat$^{\ast}$	     &0.5268&0.198&0.250	 &0.448&0.096&0.152	     &0.487&0.147&0.201\\
Llama-2-13B-Chat$^{\ast}$        &0.574&0.242&0.280	     &0.488&0.094&0.178	     &0.531&0.168&0.229\\
Llama-3-8B-Instruct$^{\ast}$     &0.656&0.300&0.356	     &0.562&0.122&0.238	     &0.609&0.211&0.297\\
Llama-3-70B-Instruct$^{\ast}$    &0.750&0.422&0.498	     &0.642&0.178&0.330	     &0.696&0.300&0.414\\
Mistral-7B-Instruct-V03$^{\ast}$ &0.227&0.072&0.086	     &0.148&0.008&0.022      &0.188&0.040&0.054\\
Conifer-Base$^{\ast}$	         &0.510&0.184&0.232	     &0.300&0.018&0.048	     &0.405&0.101&0.140\\
Conifer-Test$^{\ast}$	         &0.559&0.215&0.255	     &0.328&0.102&0.156	     &0.443&0.159&0.206\\
\midrule
BaiChuan-13B-Chat	        &0.630&0.306&0.366	&0.521&0.114&0.196	&0.575&0.210&0.281\\
BaiChuan2-13B-Chat	        &0.669&0.348&0.418	&0.547&0.134&0.226	&0.608&0.241&0.322\\
Llama-3-8B-Instruct-CN      &0.743&0.458&0.510	&0.627&0.162&0.314	&0.685&0.310&0.412\\
Llama-3-70B-Instruct-CN     &0.756&0.482&0.536	&0.636&0.190&0.322	&0.696&0.336&0.429\\
DeepSeek-7B-Chat            &0.695&0.378&0.442	&0.580&0.150&0.270	&0.638&0.264&0.356\\
DeepSeek-V2-Lite-Chat       &0.733&0.382&0.448	&0.597&0.148&0.262	&0.665&0.265&0.355\\
DeepSeek-67B-Chat           &0.802&0.516&0.578	&0.662&0.180&0.350	&0.732&0.348&0.464\\
InternLM2-Chat-7B           &0.767&0.452&0.538	&0.625&0.172&0.320	&0.696&0.312&0.429\\
GLM-4-9B-Chat               &0.885&0.678&0.742	&0.742&0.288&0.450	&0.813&0.483&0.596\\
YI-1.5-34B-Chat             &0.881&0.672&0.740	&0.745&0.302&0.474	&0.813&0.487&0.607\\

%Qwen1.5_4B_chat   &0.732&0.380&0.442 &0.457&0.034&0.096 &0.594&0.207&0.269\\
% Qwen1.5_7B_chat   &0.908&0.684&0.752 &0.545&0.044&0.140 &0.726&0.364&0.446\\
Qwen1.5-4B              &0.454&0.170&0.198	&0.376&0.074&0.116	&0.415&0.122&0.157\\
Qwen1.5-4B-Chat         &0.652&0.310&0.362	&0.536&0.104&0.198	&0.594&0.207&0.280\\
Qwen1.5-7B              &0.473&0.176&0.212	&0.400&0.090&0.142	&0.437&0.133&0.177\\
Qwen1.5-7B-Chat         &0.799&0.534&0.592	&0.654&0.194&0.338	&0.726&0.364&0.465\\
Qwen1.5-14B             &0.498&0.228&0.280	&0.430&0.110&0.176	&0.464&0.169&0.228\\
Qwen1.5-14B-Chat        &0.822&0.558&0.626  &0.671&0.202&0.370	&0.746&0.380&0.498\\
Qwen1.5-32B             &0.647&0.336&0.408	&0.528&0.132&0.224	&0.587&0.234&0.316\\
Qwen1.5-32B-Chat        &0.883&0.678&0.744	&0.704&0.228&0.412	&0.793&0.453&0.578\\
Qwen1.5-72B             &0.627&0.324&0.380	&0.556&0.148&0.248	&0.591&0.236&0.314\\
Qwen1.5-72B-Chat        &0.896&0.710&0.776  &0.730&0.254&0.436	&0.813&0.482&0.606\\
Qwen1.5-110B-Chat       &0.905&0.724&0.792	&0.730&0.276&0.438	&0.818&0.500&0.615\\
Qwen2-0.5B-Instruct     &0.446&0.150&0.172	&0.393&0.070&0.110	&0.419&0.110&0.141\\
Qwen2-1.5B-Instruct     &0.607&0.250&0.316	&0.496&0.104&0.168	&0.551&0.177&0.242 \\
Qwen2-7B                &0.576&0.260&0.316	&0.478&0.120&0.192	&0.527&0.190&0.254\\
Qwen2-7B-Instruct       &0.835&0.584&0.642	&0.682&0.198&0.362	&0.758&0.391&0.502\\
Qwen2-72B               &0.711&0.424&0.484	&0.568&0.170&0.274	&0.640&0.297&0.379\\
Qwen2-72B-Instruct      &\uwave{0.944}&\uwave{0.836}&\uwave{0.880}	&\uwave{0.791}&0.342&0.530	&\uwave{0.867}&0.589&\uwave{0.705}\\
\bottomrule[0.1em]
\end{tabularx}
\caption{\label{Complete Experiments Result}
The complete evaluation results and rankings of CFBench and its respective subsets. Notably, $^{\ast}$ stands for the model supporting mainstream languages excluding Chinese, and $^{\dag}$ represents calling through the API. The \textbf{bold}, \uline{underlined}, and \uwave{tilde} denote the first, second, and third rankings, respectively. Llama-3-8B-Instruct-CN and Llama-3-70B-Instruct-CN respectively represent Llama-3-8B-Instruct-Chinese and Llama-3-70B-Instruct-Chinese, both of which have undergone Chinese SFT (Supervised Fine-Tuning). Both conifer_base and conifer_test are based on the Mistral-7B foundational model. For the base model, we used a 3-shot approach for generation.
 }
\end{table*}
%实验结果
% 

%约束体系：
\newcolumntype{C}{>{\centering\arraybackslash}X}
\begin{table*}
\centering
\footnotesize % 更小的字体

\begin{tabularx}{\textwidth}{m{2.4cm} m{2.2cm} m{7cm} m{2.9cm}}
\toprule[0.1em]
\textbf{Primary} &\textbf{Secondary} &\centering\textbf{Definition} &\hspace{30pt}\textbf{Example}\\
\midrule
\multirow{5}*{Content Constraint} & Lexical &Mandatory use of specific terms or symbols, including their inclusion and precise placement. &...must include the word "beautiful." \\
                                  & Element &Mandates for including specific elements or concepts in responses, reflecting a scenario or object. &...highlights the Great Wall.  \\
                                  & Semantic &Directives on thematic content, perspective, or tone, emphasizing response significance. &Write a poem about London.  \\
\midrule
\multirow{4}*{Numerical Constraint} & Word Count &Limit the number of words or tokens. &A 50-word poem. \\
                                  & Sentence Count &Limit the number of sentences. &... three sentences.  \\
                                  & Paragraph Count &Limit the number of paragraphs. &divided into 3 sections.  \\
                                  & Document Count &Limit the number of documents. &... list 3 articles.  \\
\midrule
\multirow{10}*{Stylistic Constraint} & Tone and emotion &The emotional tone must adhere to standards such as seriousness, anger, joy, humor, and politeness. &Write a letter in an angry and sarcastic tone. \\
                                  & Form and style &Text expression standards ensure alignment with specific stylistic criteria in both presentation and perception. &Write a passage in an encyclopedic style. \\
                                  & Audience-specific &Text should be tailored to specific audiences, ensuring clarity and relevance for children, students, or specialized groups. &Write a pome for a 6-year-old. \\
                                  & Authorial style &Texts should emulate the styles of authors like Shakespeare to achieve artistic effects or depth. &Write a passage the style of Shakespeare.  \\
\midrule
\multirow{7}*{Format Constraint} & Fundamental &Widely accepted and utilized standard formats, including JSON, XML, LaTeX, HTML, Table, and Markdown.  &Extract keywords and output in JSON format.\\
                                  & Bespoke &Protocols for information expression tailored to specific needs,including paragraphing, headings, text emphasis, examples, and bullet points.  &Summarize the main idea and output in unordered list format.\\
                                  & Specialized &Formatting standards tailored for specialized applications or  domains.  &Conform to electronic medical record format.\\
\midrule
\multirow{10}*{Linguistic Constraint} & Pragmatic &Contextual language study, encompassing speech acts, implicature, discourse, dialects, sociolects, and language policy. &Output in English, in classical Chinese style.\\
                                  & Syntactic &Sentence structure, including phrases, constituents, subordinate clauses, ba-constructions, and imperatives.  &Use imperatives with nouns and verb phrases.\\
                                  & Morphological &The internal structure and formation rules of words, including roots, affixes, and morphological changes.  &Output all content in lowercase English.\\
                                  & Phonological &Study on phonological structures:phonemes, allophones, pitch, duration, and intensity.  &Single-rhyme tongue twisters.\\
\midrule
\multirow{5}*{Situation Constraint} & Role-based &Simulating characters based on context, emulating their traits, language, and behaviors. &You are Confucius, how do you decide? \\
                                  & Task-specific &Offer tailored solutions based on a nuanced understanding of situational demands. &Must work from home, how to report?  \\
                                  & Complex context &Reasoning and problem-solving within intricate and multifaceted contexts. &4 on the left, 10 total, which from right?  \\
\midrule
\multirow{1}*{Example Constraint} &\hspace{30pt}- &Regulate new responses by leveraging intrinsic patterns from a limited set of samples. &Example:input:xxx, output:\{...\}; input:xx, output? \\

\midrule
\multirow{1}*{Inverse Constraint} &\hspace{30pt}- &Narrow the response space through inverse constraints and indirect exclusion. &Prohibited from answer political topics. \\

\midrule
\multirow{1}*{Contradictory Constraint} &\hspace{30pt}- &Mutually exclusive constraints prevent fulfilling all requirements concurrently. &Write a five-character quatrain, 1000 words.  \\

\midrule
\multirow{1}*{Rule Constraint} &\hspace{30pt}- &Standardize the road of responses through meticulously crafted logic flows or actions.   &Each answer adds 1, 1+1=3, then 2+3=?\\

\bottomrule[0.1em]
\end{tabularx}
\caption{\label{constraint system}Constraint System of CFBench}
\end{table*}
%约束体系

% 领域分布
\begin{table*}
\centering
\footnotesize % 更小的字体
\begin{tabularx}{\textwidth}{p{1.65cm}m{3.2cm}m{3.2cm}m{3.2cm}m{3.2cm}}
\toprule[0.1em]
\textbf{Domain} &\multicolumn{4}{c}{\textbf{Scenarios List}} \\

\midrule
\multirow{3}*{Healthcare}  &Symptom Consultation &Diagnostic Explanation &Medication Guidance &Procedures      \\
                        &Wellness &Medical Info &Guidelines Inquiry &Public Health       \\
                        &Medical Education & & &       \\

\multirow{3}*{Education}  &Teaching Methods &Resource Access &Curriculum Design &Communication      \\
                        &Academic Counseling &Mental Health Support &Tutoring &Subject Q\&A       \\
                        &Reports &Interests & &       \\

\multirow{4}*{Finance}  &Market Research  &Stock Analysis  &Investment Analysis  &Personal Finance      \\
                        &Corporate Tax  &Insurance Management &Corporate Financing &Compliance \& Risk        \\
                        &Financial Education  &Product Development &Customer Service &Financial Reports        \\
                        &Regulatory Analysis & & &       \\

\multirow{3}*{Legal}  &Legal Education  &Legal Consultation  &Document Review  &Case Analysis      \\
                        &Statute Explanation  &Regulation Analysis  &IP Management  &Legal Training        \\
                        &Case Management  &Compliance \& Risk  &  &         \\
\multirow{1}*{Media}  &Content Creation  &Information Analysis  &Marketing \& Promotion  &News Reporting      \\
\multirow{1}*{Tourism}  &Travel Consultation  &Itinerary Planning  &Route Introduction  &      \\
\multirow{3}*{Recruitment}  &JD Writing \& Analysis  &Resume Creation  &Resume Screening  &Interview Preparation      \\
                        &Interview Evaluation  &Career Planning  &Offer Comparison  &Communication Skills        \\
                        &Performance Review  &  &  &         \\
\multirow{3}*{Gov Affairs}  &Policy Research  &Public Education  &Service Guide  &Public Services      \\
                        &Document Writing  &Content Review  &Business Procedures  &Civil Servant Training        \\
                        &Emergency Management  &  &  &         \\
\multirow{4}*{Real Estate}  &Purchase Planning  &Market Trends  &Property Policies  &Development      \\
                        &Leasing  &Property Valuation  &Amenities  &Financial Services        \\
                        &Property Description  &Content Creation  &Sales \& Marketing  &Qualifications         \\
                        &Renovation  & & &       \\
\multirow{3}*{Automotive}  &Marketing \& Sales  &Driving \& Safety  &Customer Experience  &Model Consultation      \\
                        &Model Comparison  &Loan Calculation  &Insurance Evaluation  &Claims Assessment       \\
                        &Car Reviews  &Maintenance \& Repair  &  &         \\
\multirow{3}*{Psychology}  &Romance  &Family  &Friendship  &Workplace      \\
                        &Self \& Health  &Social  &Sexuality \& Gender  &Life Stages        \\
                        &Organizational  &Client Relations  &Crisis Intervention  &Public Psychology         \\
\multirow{3}*{Internet}  &Business Analysis  &Product Design  &User Research  &Coding \& Debugging      \\
                        &Product Testing  &Data Management  &Cybersecurity  &Computer Q\&A        \\
                        &Internet News  &Marketing  &Operations  &UI/UX Design         \\
\multirow{2}*{Spirituality}  &Beliefs \& Rituals  &Divination  &Feng Shui  &Astrology      \\
                        &Metaphysics  &Spirituality  &Healing  &Content Review        \\
\multirow{3}*{Sports}  &Training  &Goal Setting  &Nutrition  &Workout Plans      \\
                        &Equipment \& Tech  &Performance  &Injury Care  &Specialized Training        \\
                        &Mental Motivation  &Data Tracking  &  &         \\
\multirow{3}*{Lifestyle}  &Life Tips  &Shopping Decisions  &Instant Queries  &Skincare      \\
                        &Fashion \& Styling  &Naming  &Recommendations  &Planning        \\
                        &Socializing  &Life Creations  &Q\&A  &         \\
\multirow{3}*{Culturtainment}  &Podcasts \& Radio  &TV \& Film  &Music  &Literature      \\
                        &Theater \& Dance  &Art  &Cultural Events  &Short Videos \& Live        \\
                        &Gossip  &Content Creation  &  &         \\
\multirow{2}*{Employment}  &Project Management  &Translation  &Office Efficiency  &Marketing      \\
                        &Administration  &Customer Service  &Team Collaboration  &Collaboration        \\
\multirow{3}*{Cuisine}  &Food \& Restaurant Recs  &Reviews \& Feedback  &Marketing \& Promo  &Food Content      \\
                        &Culinary Culture  &Recipes \& Menus  &Cooking Techniques  &Ingredient Prep        \\
                        &Nutrition \& Health  &Food Safety  &Culinary Training  &         \\
\multirow{3}*{Gaming}  &Guides  &Reviews  &Hardware \& Peripherals  &News      \\
                        &Software \& Services  &Development \& Design  &Operations  &Mini Games        \\
                        &Search  &Marketing \& Promotion  &Esports \& Tournaments  &Culture \& Education         \\
\multirow{2}*{Technology}  &Reviews  &Launches  &Buying Guides  &Tips \& Tricks      \\
                        &Content Creation  &Product Design  &Marketing Copy  &After-sales \& Repairs        \\

\bottomrule[0.1em]
\end{tabularx}
\caption{\label{Domain and Scenarios List}Domain and Scenarios List
}
\end{table*}

\iffalse
\begin{table}
\centering
\begin{tabularx}{\linewidth}{p{3cm}ccc}
\toprule[0.1em]
\textbf{Models} &\textbf{Easy Set} &\textbf{Hard Set} &\textbf{All Set} \\
\midrule
llama2_7B  &60 &55 &58   \\
llama2_13B  &64 &61 &63   \\
llama2_70B  &71 &69 &70    \\
llama2_13B_chat  &65 &62 &63    \\
llama2_13B_chat v1  &66 &63 &64    \\
\midrule
conifer_base  &63 &62 &62.5    \\
conifer  &65 &63 &64    \\
\midrule
toSimple_base  &62 &61 &61.5    \\
toSimple_SFT  &63 &62 &62.6    \\
toSimple_DPO  &65 &64.2 &64.5    \\
\bottomrule[0.1em]
\end{tabularx}
\caption{\label{improve ability} The way of improve follow ability.
}
\end{table}
\fi

%Benchmark对比：
\newcolumntype{C}{>{\centering\arraybackslash}X}
\begin{table*}

\footnotesize % 小号字体
\begin{tabularx}{\textwidth}{>{\centering\arraybackslash}p{1.5cm} m{5.1cm} m{3.7cm} m{4cm}}
\toprule[0.1em]
\multicolumn{1}{c}{\centering\textbf{Benchmark}} & \textbf{Case} & \textbf{Features} & \textbf{{CFBench's Advantages}} \\
\midrule

\textbf{Complex \newline Bench} 
& \textbf{Instruction:} Please introduce the following painting. 
\newline - If the work contains any animal, the description should be in English. 
\newline - Otherwise, the description should be in Chinese. 
\newline \textbf{Response:} Painting: "Mona Lisa" 
& 
Focus on the \textbf{combination} of instructions, including And/Chain/Selection/Nested.
\newline
\includegraphics[width=0.5\linewidth]{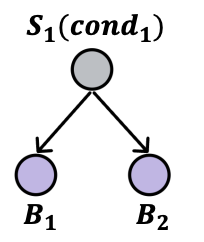}
& 
CFBench supports \textbf{simplicity in instruction compliance}, \textbf{broad applicability} across use cases, and \textbf{scalability} for various tasks. 

\\ \hline

\textbf{Follow \newline Bench} 
& \textbf{Instruction:} Using following words to generate a story. 
\newline "needed, money, computer, bought, happy". 
\newline \textbf{Response:} In the heart of the city, there was a research facility that desperately needed funding for a crucial project. To gather the required money, ··· 
& 
Focus on \textbf{gradual constraint escalation}. The more constraints per instruction, the greater the complexity.
\newline
\includegraphics[width=0.5\linewidth]{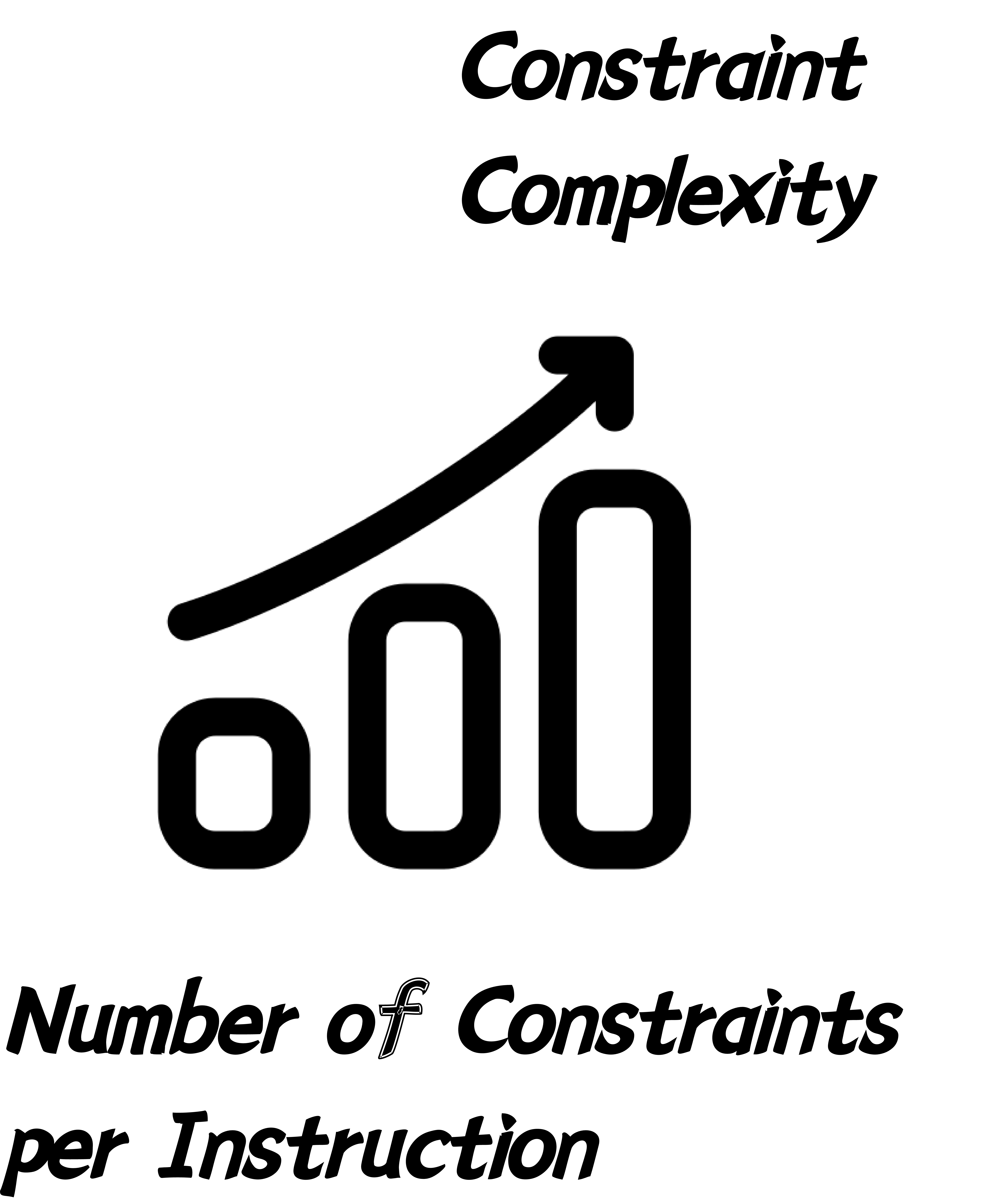}
& 
CFBench is \textbf{simple} to use, \textbf{resource-efficient}, and \textbf{scalable} for large models.

\\ \hline

\textbf{IFEval} 
& \textbf{Instruction:} Given the sentence "Two young boys with toy guns and horns." can you ask a question? Please ensure that your response is in English, and in all lowercase letters. No capital letters are allowed. 
\newline \textbf{Response:} what are the two young boys holding? 
& 
\textbf{Verifiable Instructions}.
\newline

{\centering
\includegraphics[width=0.4\linewidth]{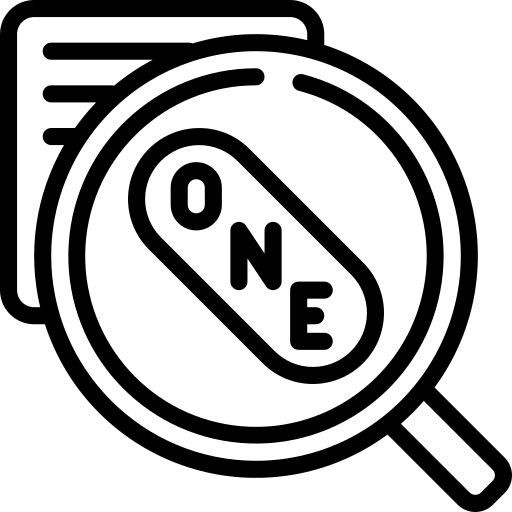}\centering
}& 
CFBench is \textbf{easy to implement}, \textbf{general-purpose across models}, and \textbf{scalable for large datasets}.
 
\\ \hline

\textbf{CFBench} 
& \textbf{Instruction:} Overview of the Trump Shooting Incident, include the cause, process, aftermath, shooter information and Trump's feedback. Organize the content using numbered points and bold text. The cause should be in JSON format with event keywords: time, location, and nature. Summarize the process in three sentences. Discuss the impact from both international and domestic perspectives, keeping the total content under 500 words. Finally, summarize this event in a seven-word quatrain. 
\newline \textbf{Response:} 1. Cause \textbackslash n{"time":"July 13, 2024", "location":"Butler, Pennsylvania, USA", "nature":"assassination attempt"} \textbackslash n 2. Process \textbackslash n-Trump was giving a speech at the rally when the shooting occurred. \textbackslash n-Secret Service agents quickly surrounded Trump and escorted him away. \textbackslash n 3. ···

& 
\textbf{Comprehensive Coverage}: 
\newline Multi-dimensional constraints; 
\newline Cross-task evaluation; 
\newline Real-world complexity.
\newline
\textbf{Standardized Framework}: 
\newline Unified metrics; 
\newline Cross-model compatibility.
\newline
\includegraphics[width=1\linewidth]{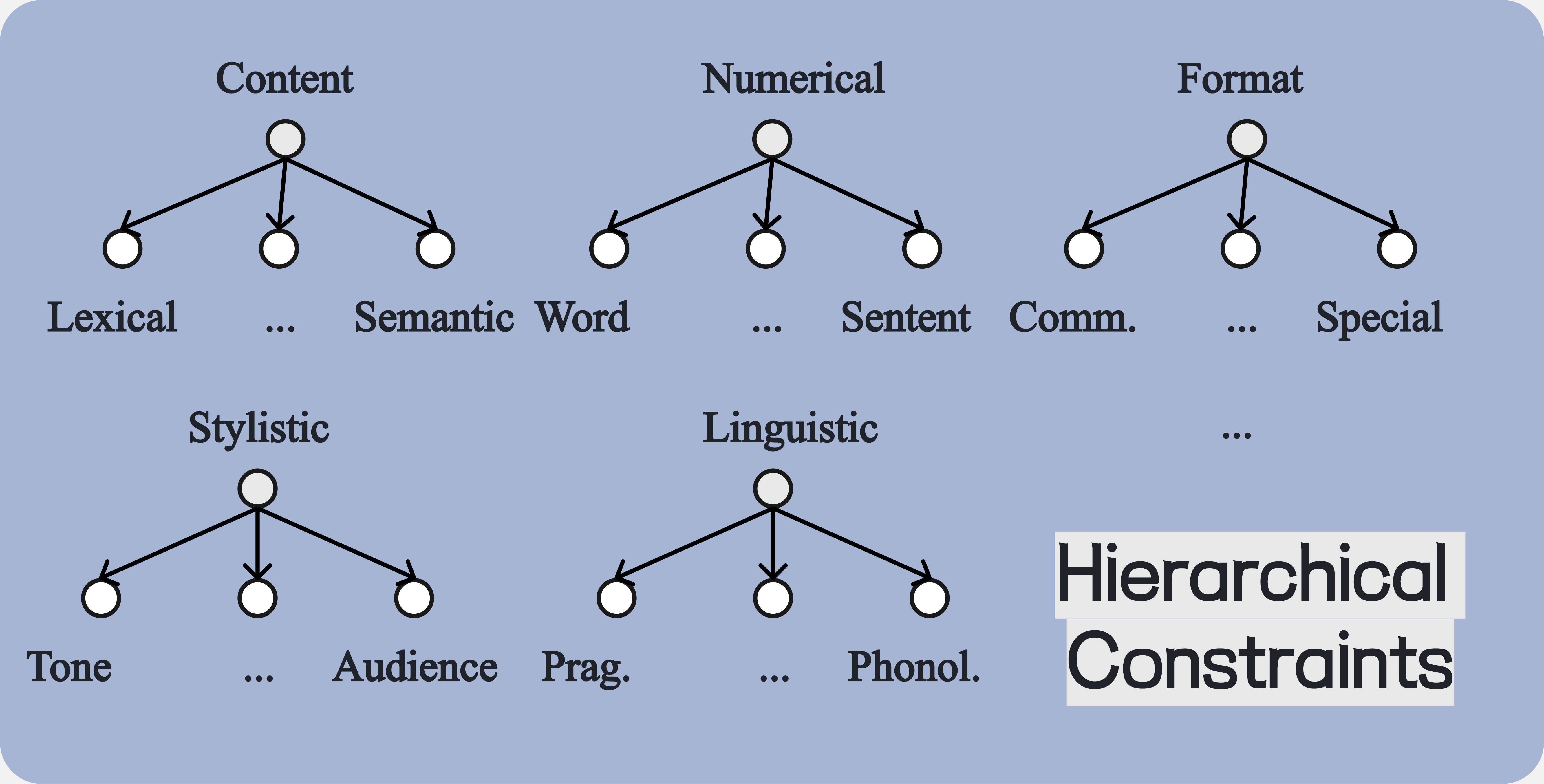}
\includegraphics[width=1\linewidth]{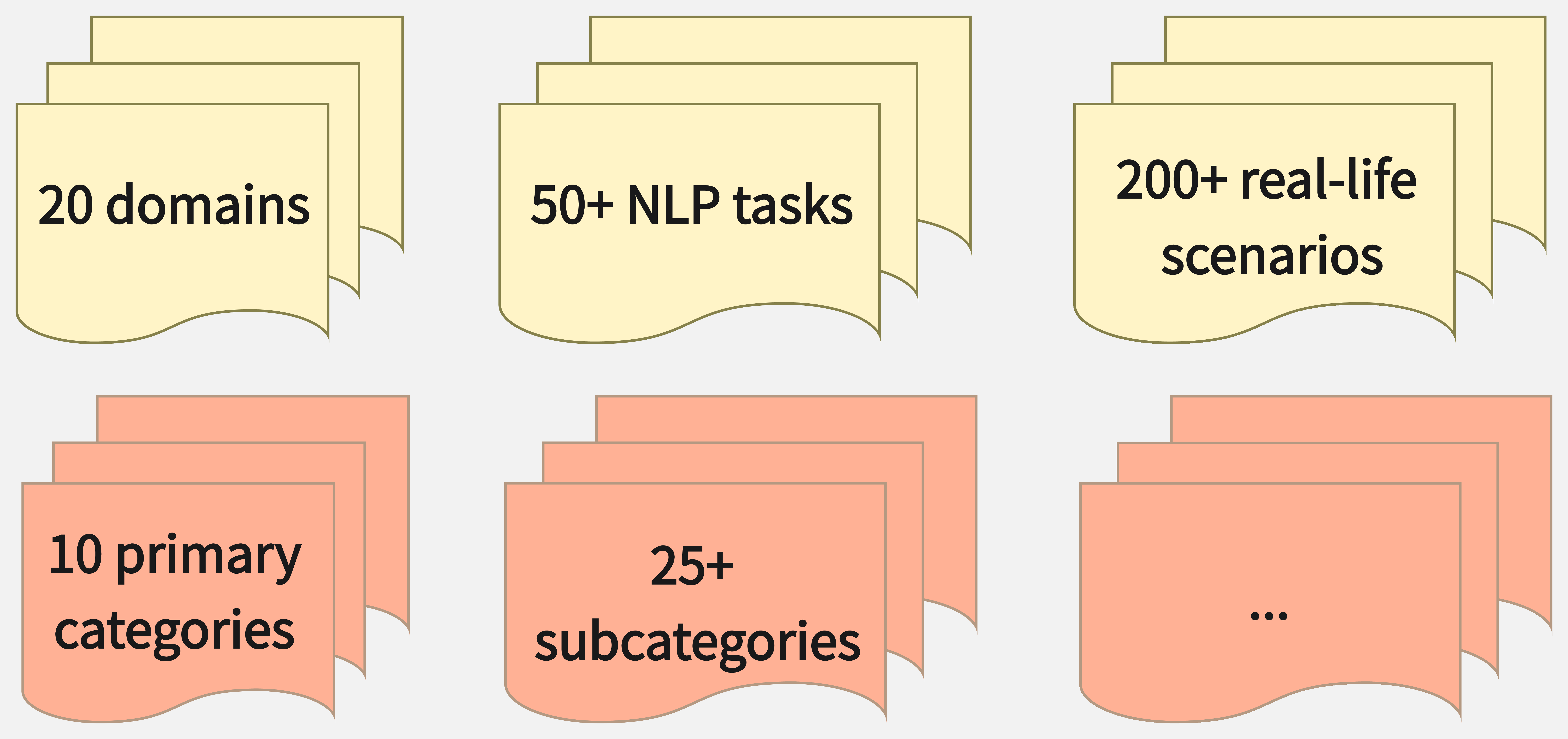}
& 
\textbf{Diverse instruction types}, \newline
\textbf{standardized evaluation}, \newline
\textbf{hierarchical constraints}.

\\ 
\bottomrule[0.1em]
\end{tabularx}
\caption{\label{Benchmark Comparison} Benchmark Comparison for Different Instruction Types and Task Complexity}
\end{table*}

%Benchmark对比

% 图:
% 图:

\begin{figure*}[t]
\centering
\includegraphics[width=0.9\textwidth]{cfbench_fig/11_case_en_cn.pdf}  
\caption{CFBench Example: English-Chinese Comparison, Data Itself in Chinese.} 
\label{CFBench Case: Chinese and English Comparison} 
\end{figure*}

% 图:
\end{document}